\pdfoutput=1
\documentclass{article}

\usepackage{arxiv}
\usepackage{latexml}

\usepackage[utf8]{inputenc}
\usepackage[T1]{fontenc}
\usepackage{times}
\usepackage[authoryear,round]{natbib}
\setcitestyle{citesep={;},aysep={,},yysep={;}}

\usepackage{amsmath,amsfonts,bm}

\def\eqref#1{equation~\ref{#1}}

\def\1{\bm{1}}

\DeclareMathAlphabet{\mathsfit}{\encodingdefault}{\sfdefault}{m}{sl}
\SetMathAlphabet{\mathsfit}{bold}{\encodingdefault}{\sfdefault}{bx}{n}

\usepackage{hyperref}
\usepackage{url}
\usepackage{booktabs}
\usepackage{graphicx}
\usepackage{multirow}
\usepackage{multicol}
\usepackage{wrapfig}
\usepackage{xcolor}

\title{StreamEgo: Streaming Exocentric-to-Egocentric Video Generation}

\renewcommand{\shorttitle}{StreamEgo: Streaming Exocentric-to-Egocentric Video Generation}

\iflatexml
\author{Tianyi Xu\thanks{Equal contribution.}\\
Peking University\\
AI\textsuperscript{2} Robotics\\
\texttt{photon@stu.pku.edu.cn}
\and
Boyang Lyu\thanks{Equal contribution.}\\
Peking University\\
AI\textsuperscript{2} Robotics
\and
Yunkai Tang\\
Peking University\\
AI\textsuperscript{2} Robotics
\and
Kairan Ding\\
Peking University\\
AI\textsuperscript{2} Robotics
\and
Jiahang Li\\
Peking University\\
AI\textsuperscript{2} Robotics
\and
Weiyu Guo\thanks{Project leader.}\\
AI\textsuperscript{2} Robotics
\and
Yandong Guo\\
AI\textsuperscript{2} Robotics
\and
Shuchen Weng\thanks{Corresponding author.}\\
Beijing Academy of Artificial Intelligence (BAAI)\\
Peking University\\
\texttt{shuchenweng.cn@gmail.com}
\and
Boxin Shi\thanks{Corresponding author.}\\
Peking University\\
\texttt{shiboxin@pku.edu.cn}
}
\else
\author{%
\textbf{Tianyi Xu}$^{1,2,*}$\quad
\textbf{Boyang Lyu}$^{1,2,*}$\quad
\textbf{Yunkai Tang}$^{1,2}$\quad
\textbf{Kairan Ding}$^{1,2}$\quad
\textbf{Jiahang Li}$^{1,2}$\\
\textbf{Weiyu Guo}$^{2,\#}$\quad
\textbf{Yandong Guo}$^{2}$\quad
\textbf{Shuchen Weng}$^{3,1,\dagger}$\quad
\textbf{Boxin Shi}$^{1,\dagger}$\\[0.6em]
\normalfont $^{1}$Peking University\quad
$^{2}$AI$^{2}$ Robotics\quad
$^{3}$Beijing Academy of Artificial Intelligence (BAAI)\\[0.2em]
\normalfont\small\texttt{photon@stu.pku.edu.cn}\quad
\texttt{shuchenweng.cn@gmail.com}\quad
\texttt{shiboxin@pku.edu.cn}\\[0.4em]
\normalfont\footnotesize $^{*}$Equal contribution\quad
$^{\#}$Project leader\quad
$^{\dagger}$Corresponding authors
}
\fi

\date{}

\begin{document}

\maketitle

\begin{abstract}
Exocentric-to-egocentric (Exo2Ego) video generation transforms third-person recordings into first-person viewpoints, unlocking immersive AR/VR experiences and scalable demonstration synthesis for embodied robotics. Extending it to a streaming setting is expected to further enable live applications. However, existing methods suffer from degraded manipulation regions, overfitting and confinement to fixed-length clips with prohibitive latency. To address these limitations, we present StreamEgo, a real-time framework for unbounded, generalizable Exo2Ego video generation. To resolve intricate manipulation dynamics while supporting the streaming pipeline, a streaming dual-branch prior provides pixel-aligned guidance for translating appearance and skeletal kinematics. We also introduce a region-aware loss curriculum to improve generalization and stabilize multi-objective diffusion training. Furthermore, we distill the model into a chunk-wise autoregressive model, augmented with a region-aware corruption mechanism and a first-chunk curriculum, enabling high-quality, generalizable real-time streaming Exo2Ego generation at 24.3 FPS on a single H100 GPU. %
\end{abstract}

\section{Introduction}
\label{sec:intro}

Exocentric-to-egocentric (Exo2Ego) video generation~\citep{liu2024exocentric,luo2024put,kang2025egoxegocentricvideogeneration,mahdi2025exo2egosyn,xu2025egoexogen,wang2026groundedexo2ego,mahdi2026synchronysequenceexotoegogeneration} converts third-person recordings into dynamic first-person viewpoints. For embodied AI, Exo2Ego unlocks massive web-scale video repositories as a scalable and diverse data source. Real-time streaming Exo2Ego further extends this capability to live applications such as AR-based tutorials and surgical demonstrations. %

\begin{figure}[tbp]
\centering
\includegraphics[width=\textwidth]{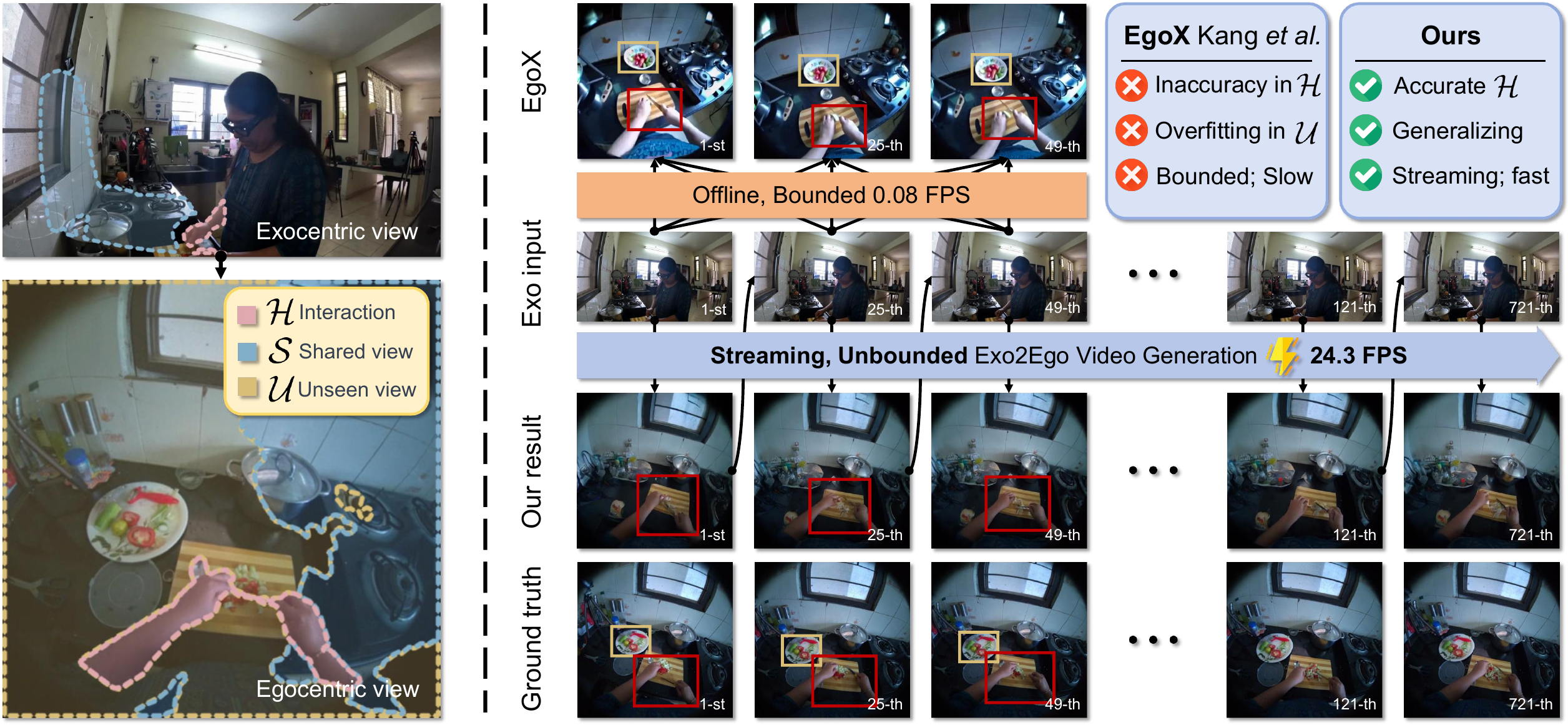}
\caption{Streaming exocentric-to-egocentric (Exo2Ego) video generation. \textbf{Left:} Tri-regional decomposition of the task of Exo2Ego. \textbf{Right:} Comparison of our method with state-of-the-art Exo2Ego method EgoX~\citep{kang2025egoxegocentricvideogeneration} on Ego-Exo4D dataset~\citep{grauman2024egoexo4d}.}
\label{fig:teaser}
\end{figure}

To achieve accurate Exo2Ego video generation, recent studies~\citep{kang2025egoxegocentricvideogeneration,xu2025egoexogen} formulate Exo2Ego by decomposing the egocentric canvas into three functional regimes (Fig.~\ref{fig:teaser}:\textbf{Left}): the hand-object interaction area ($\mathcal{H}$), the shared view ($\mathcal{S}$), and the unseen view ($\mathcal{U}$). %
By leveraging geometric reprojections from the exocentric view to construct appearance priors, existing methods~\citep{kang2025egoxegocentricvideogeneration,yin2026camdirector,wang2026groundedexo2ego} achieve robust alignment and satisfactory visual fidelity in $\mathcal{S}$.

However, they fall short across the remaining domains. 
Firstly, accurate hand-object interaction remains challenging in $\mathcal{H}$, as projective geometric priors break down under complex foreground dynamics, while existing mask-~\citep{xu2025egoexogen} or pose-based~\citep{park2026egoworld} remedies are fragile due to their reliance on ground-truth first-frame or noisy hand pose estimates (Fig.~\ref{fig:teaser}:\textbf{Right}, red box).
Meanwhile, existing approaches suffer from benchmark memorization on paired datasets~\citep{grauman2024egoexo4d}, artificially replicating ground-truth details in $\mathcal{U}$ and impairing generalization (Fig.~\ref{fig:teaser}:\textbf{Right}, %
yellow box).
Finally, existing methods rely on full-sequence offline scanning for appearance priors~\citep{yin2026camdirector,kang2025egoxegocentricvideogeneration,wang2026groundedexo2ego} and full-sequence bidirectional attention, implicitly assuming access to future frames. This non-causal paradigm is fundamentally incompatible with streaming, where the future is unknown (Fig.~\ref{fig:teaser}:\textbf{Right}, 0.08 FPS).

In this paper, we propose StreamEgo, a real-time, streaming, and generalizable framework for Exo2Ego video generation.
To resolve fine-grained manipulation in $\mathcal{H}$ under the streaming setting, we establish a pixel-aligned conditioning scheme via streaming dual-branch priors. A dynamic ego prior maintains an incremental 3D world cache for the static background and tracks moving entities to render ego-aligned appearance anchors, while a human pose prior provides structural guidance for hand interactions.
To improve the quality in $\mathcal{U}$, we propose a region-aware training curriculum. It progressively shifts the optimization focus from $\mathcal{S}$ to $\mathcal{U}$ and then to $\mathcal{H}$. This curriculum stabilizes training while improving generalization.
To support streaming under a causal setting, we adapt a chunk-wise autoregressive architecture, where each chunk is generated only from past chunks, and distill the diffusion backbone into a high-throughput causal student with $3$ steps. This removes the reliance on full-sequence offline scanning and bidirectional attention. Meanwhile, we propose a region-aware corruption mechanism and a first-chunk curriculum to improve the quality for the streaming rollouts.
As a result, StreamEgo can process unbounded exocentric video streams and generate high-quality egocentric videos at 24.3 FPS on a single H100 GPU.

Our main contributions are summarized as follows:
\begin{itemize}
    \item We present StreamEgo, the first streaming framework for Exo2Ego video generation, reformulating the task from full-sequence generation to causal generation.
    \item We introduce a streaming dual-branch prior and a region-aware training curriculum, which respectively improve interaction fidelity in $\mathcal{H}$ and generalization in $\mathcal{U}$.
    \item We introduce a region-aware corruption mechanism and a first-chunk curriculum, enhancing the visual quality for StreamEgo, achieving high-quality real-time Exo2Ego generation.
\end{itemize}

\section{Related Work}
\label{sec:relatedwork}

\subsection{Exocentric-to-Egocentric Video Generation}
\label{sec:related_exo_to_ego}

Paired Exo-Ego datasets~\citep{grauman2024egoexo4d,h2o,sener2022assembly101} have catalyzed exocentric-to-egocentric synthesis. Early attempts simplify the setup via multi-view inputs~\citep{liu2024exocentric,mahdi2025exo2egosyn}, initial ego guidance~\citep{xu2025egoexogen}, or static image translation~\citep{park2026egoworld}. Concurrently, novel-view synthesis methods inject camera encodings~\citep{bai2025recammaster,bai2024syncammaster,li2025prope,zhang2026ucpe,yu2025trajectorycrafter} or condition on rendered 3D/4D geometry~\citep{ren2025gen3c,cao2025uni3c,bahmani2026lyra,lin2026vista4d}, but falter under extreme perspective shifts. Structured conditions, such as sketches for video instance repainting~\citep{weng2025vires}, multi-stream audio cues for temporal control~\citep{weng2025audiosync}, and latitude/longitude-aware mechanisms for 360$^\circ$ generation~\citep{xia2025panowan}, have also proven effective in steering video diffusion models toward precise spatial and temporal control. Recent single-exo works incorporate geometry guidance~\citep{kang2025egoxegocentricvideogeneration} or semantic grounding~\citep{wang2026groundedexo2ego}, yet uniform canvas supervision causes benchmark memorization and hand drift across short, fixed-length clips. In contrast, StreamEgo addresses these problems with streaming dual-branch priors and region-aware curriculum scheduling, enabling generalizable, real-time streaming synthesis.

\subsection{Streaming Video Generation Models}
\label{sec:related_autoregressive}

To break the temporal limits of short-clip diffusion, training-free noise rescheduling~\citep{qiu2023freenoise,kim2024fifodiffusion} and training-based autoregressive conditioning~\citep{chen2024diffusionforcing,ruhe2024rollingdiffusion,sun2025ardiffusion,song2025historyguided,huang2025selfforcing} have been proposed to mitigate exposure bias. Subsequent frameworks scale temporal length via predictive modelling, long-context architectures, and causal forcing~\citep{zhang2025framepack,gu2025longcontext,liu2025infinitystar,li2025stablevideoinfinity,liu2025rollingforcing,zhu2026causalforcing}. Streaming generation has also been extended to joint audio-video synthesis, now with a dedicated benchmark~\citep{liu2026streamavbench}, and to panoramic world models for language-guided navigation~\citep{liu2026spwnav}. However, these methods focus on text-, audio-, or action-conditioned generation rather than asymmetric cross-view streaming. Built upon streaming long-video designs and few-step distillation paradigms~\citep{yuan2026helios,zhao2023unipc,yin2024onestep,yin2024improved,yin2025causvid}, StreamEgo achieves chunk-wise autoregressive modelling for the Exo2Ego task.

\subsection{Video World Models for Embodied Learning}
\label{sec:related_embodied_world_models}

Video world models facilitate embodied learning via interactive simulation~\citep{yang2023unisim}, recently extended to multiple agents~\citep{cai2026mass,yang2026pawa}, visual planning~\citep{du2023unipi,zhou2024robodreamer}, and synthetic demonstrations~\citep{bharadhwaj2024gen2act}. Subsequent methods couple video with action prediction~\citep{li2025uva,zhu2025uwm,yan2026deltawam,li2026languageforcing}, adapt video priors via inverse dynamics~\citep{feng2025vidar}, or incorporate causal history for online control~\citep{li2026lingbotva}. In parallel, vision-language-action (VLA) models are being developed on modular open-source codebases~\citep{starvla2026} and extended with spatial memory~\citep{li2026spatialmemory} and fast reflexive control~\citep{guo2026neurovla}; like world models, they rely on large-scale embodied demonstrations. Although cross-view conversion can furnish training data for egocentric world models~\citep{tran2026egoexowm}, synthesizing continuous, structurally faithful streams remains challenging. StreamEgo bridges this gap by streaming third-person observations into structurally consistent egocentric trajectories, providing scalable first-person training data for embodied AI.

\section{Method}
\label{sec:method}

Given an input exocentric video stream along with known exocentric and target egocentric camera intrinsics, and the target egocentric camera trajectory, the goal for StreamEgo is to synthesize the corresponding egocentric video from the target viewpoint with the same frame rate as the input.

We develop StreamEgo in three stages: formulating streaming priors (Sec.~\ref{sec:dual_condition}), building the base system with the priors (Sec.~\ref{sec:method_teacher}), and making the system real-time (Sec.~\ref{sec:method_student}).

\subsection{Streaming Dual-Branch Priors}
\label{sec:dual_condition}

Exo2Ego is intrinsically underconstrained because of extreme perspective disparities. This challenge is worse in $\mathcal{H}$, where motion is rapid and occlusions are frequent. The model therefore benefits from pixel-aligned conditions that provide visual context and structural guidance. At the same time, practical online deployment requires these conditions to be built causally and in a streaming manner. To this end, we introduce streaming dual-branch priors (Fig.~\ref{fig:pipeline-teacher}\textbf{a}): (1) a streaming dynamic ego prior for appearance cues, paired with (2) a streaming human pose prior for structural guidance.

\noindent \textbf{Streaming dynamic ego prior.}
Unlike existing methods~\citep{kang2025egoxegocentricvideogeneration,yin2026camdirector} that rely on offline, full-video inputs with prohibitive latency, we introduce a streaming geometric conditioning scheme to construct causal pixel-aligned visual priors in real-time.

As is shown in Fig.~\ref{fig:pipeline-teacher}\textbf{a}, we decouple the environment into persistent static background surfaces and transient interaction geometry, following CamDirector~\citep{yin2026camdirector}. Using the stationary exocentric camera, each rectified pixel corresponds to a fixed physical ray in world space. We maintain an incremental, ray-indexed 3D world cache that stores the first valid depth and color observed along each ray. The cache updates at sparse keyframes. MoGe-2~\citep{wang2025moge2} predicts metric depth, while SAM3.1~\citep{meta2026sam31} segments dynamic entities to isolate the interaction region. Static rays outside dynamic masks are committed to the cache and remain fixed once initialized. Depth is propagated via backward optical flow between keyframes.

At frame $t$, the retrieved static background points $ \mathcal{A}_t$ and the current dynamic foreground points $ \mathcal{D}_t$ are unified in world space. The ego-aligned prior $P_t$ is synthesized via:
\begin{equation}
P_t = \mathcal{R}_{\mathcal{C}_{g,t}} \left( T_{g,t}^{-1} (\mathcal{A}_t \cup \mathcal{D}_t) \right),
\label{eq:dynamic_render}
\end{equation}
where $T_{g,t}$ is the transformation from target egocentric coordinates to world space, and $\mathcal{R}_{\mathcal{C}_{g,t}}$ is a calibrated splatting renderer parameterized by target camera intrinsics $\mathcal{C}_{g,t}$. This formulation provides stable anchors for region $\mathcal{S}$ and dynamic visual cues for contacts in $\mathcal{H}$ in a streaming fashion.

\noindent \textbf{Streaming human pose prior.}
While dynamic ego priors supply visual context, accurately synthesizing complex interactions in $\mathcal{H}$ requires explicit structural guidance~\citep{park2026egoworld}. Prior attempts either rely on ground-truth egocentric first frames for interaction masks~\citep{xu2025egoexogen}, or condition directly on isolated hand poses~\citep{park2026egoworld}; the latter induces severe spatial misalignment on the egocentric canvas because monocular depth estimation on small, dynamic hand regions is highly ill-posed. Inspired by human proprioception, we introduce skeletal constraints by kinematically bonding hand estimation to the upper-body pose.

At sampled exocentric frames, SMPLer-X~\citep{cai2023smpler} estimates and ViTPose~\citep{xu2022vitpose} refines full-body topology and upper-limb joints. Wrist joints derived from the upper-body chain define zoom-in bounding boxes for WiLoR~\citep{potamias2025wilor} to recover wrist-relative finger articulation. For each hand $s \in \{\mathrm{L}, \mathrm{R}\}$ (left and right) at frame $t$, the predicted $j$-th local finger joint $\widehat{J}^{\,s}_{t,j}$ is kinematically anchored to the refined 3D wrist joint position $B^{s}_{t,\mathrm{wrist}}$:
\begin{equation}
J^{s}_{t,j} = B^{s}_{t,\mathrm{wrist}} + \left( \widehat{J}^{\,s}_{t,j} - \widehat{J}^{\,s}_{t,\mathrm{root}} \right), \qquad s \in \{\mathrm{L}, \mathrm{R}\},
\label{eq:body_guided_hand}
\end{equation}
where $\widehat{J}^{\,s}_{t,\mathrm{root}}$ is the canonical root position of the predicted hand model. This locks the hand's global coordinate to the physical arm chain $B^{s}_{t,\mathrm{wrist}}$ while preserving local finger articulation. $B^{s}_{t,\mathrm{wrist}}$ is obtained in the calibrated exocentric camera frame by anchoring the root-relative 3D skeleton of SMPLer-X~\citep{cai2023smpler} to the known 3D position of the egocentric camera $c_t$ (Eq.~\ref{eq:dynamic_render}) and fitting its metric scale via 2D reprojection against the ViTPose~\citep{xu2022vitpose} keypoints (Appendix~\ref{sec:supp_priors}). The resulting upper-limb skeletons are causally interpolated and rasterized into both exocentric and egocentric views as skeletal guidance maps $(Q^e_t, Q^g_t)$ using calibrated camera parameters $\mathcal{C}$, with joint confidence scores modulating stroke intensities to suppress hallucinated joints.

\subsection{Bidirectional Teacher}
\label{sec:method_teacher}

\begin{figure}[tbp]
\centering
\includegraphics[width=\textwidth]{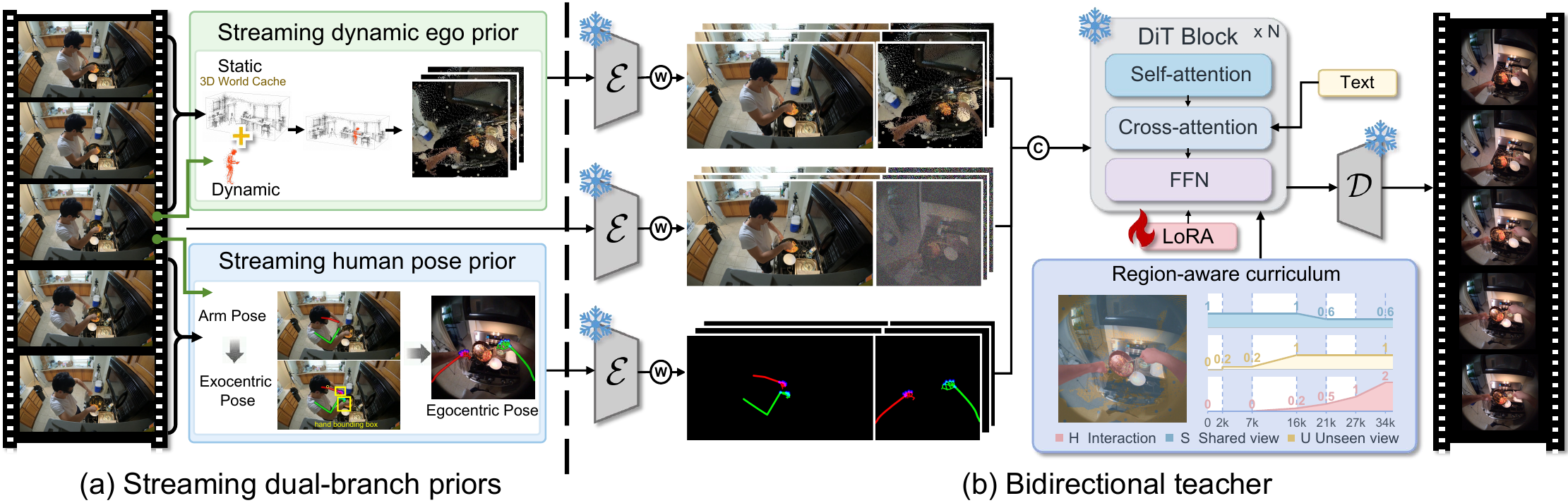}
\caption{Overview of the bidirectional teacher. \textbf{(a)} Causal construction of the streaming dual-branch priors, comprising the dynamic ego prior and human pose prior. 
\textbf{(b)} Conditioning assembly and fine-tuning of the base Wan model~\citep{team2025wan} under our region-aware curriculum.}
\label{fig:pipeline-teacher}
\end{figure}

With the streaming dual-branch priors established, we construct the teacher model with bidirectional attention to master Exo2Ego~\citep{liu2024exocentric,luo2024put,kang2025egoxegocentricvideogeneration,mahdi2025exo2egosyn,xu2025egoexogen,wang2026groundedexo2ego,mahdi2026synchronysequenceexotoegogeneration}.
As illustrated in Fig.~\ref{fig:pipeline-teacher}\textbf{b}, all visual inputs and condition priors are encoded into latent representations via a frozen video VAE.
Following EgoX~\citep{kang2025egoxegocentricvideogeneration}, we concatenate the streaming priors with video latents along the channel dimension for pixel-aligned conditioning, while concatenating the clean exocentric video and the noisy egocentric latent along the width dimension.
The composite latents are then fed into a pretrained Wan backbone. The model is trained using LoRA~\citep{hu2021lora} with our proposed region-aware training curriculum. 

\noindent \textbf{Region-aware loss weight scheduling.}
Uniform supervision across the egocentric canvas is suboptimal due to disparate learning difficulties across functional domains: $\mathcal{S}$ offers stable visual anchors, $\mathcal{U}$ requires plausible completion vulnerable to overfitting, and $\mathcal{H}$ exhibits complex dynamics that destabilize early training. We therefore propose a region-aware curriculum (Fig.~\ref{fig:pipeline-teacher}\textbf{b}): first anchoring cross-view alignment in $\mathcal{S}$, gradually introducing completion in $\mathcal{U}$, and finally ramping up supervision in $\mathcal{H}$ to refine manipulation fidelity.

Let $r_{\mathcal{S}}(p), r_{\mathcal{U}}(p), r_{\mathcal{H}}(p) \in \{0, 1\}$ denote binary regional indicators on the egocentric latent grid $\Omega_g$. At training progress $u \in [0, 1]$, the unnormalized spatial weight at position $p$ is linearly combined as $a_u(p) = \sum_{r \in \{\mathcal{S}, \mathcal{U}, \mathcal{H}\}} w_r(u) r_r(p)$, where $w_r(u)$ follows the region-aware curriculum. To stabilize gradient scales, we normalize $a_u(p)$ across the canvas into $\omega_u(p) = a_u(p) / (\frac{1}{|\Omega_g|} \sum_{q \in \Omega_g} a_u(q) + \delta)$ ($\delta > 0$). Denoting the flow-matching prediction error as $e_\sigma(p)$, our training objective is:
\begin{equation}
\mathcal{L} = \mathbb{E} \left[ \frac{1}{|\Omega_g|} \sum_{p \in \Omega_g} \omega_u(p) e_\sigma(p) \right].
\label{eq:normalized_loss}
\end{equation}

\subsection{Real-Time Autoregression}
\label{sec:method_student}

\begin{figure}[tbp]
\centering
\includegraphics[width=\textwidth]{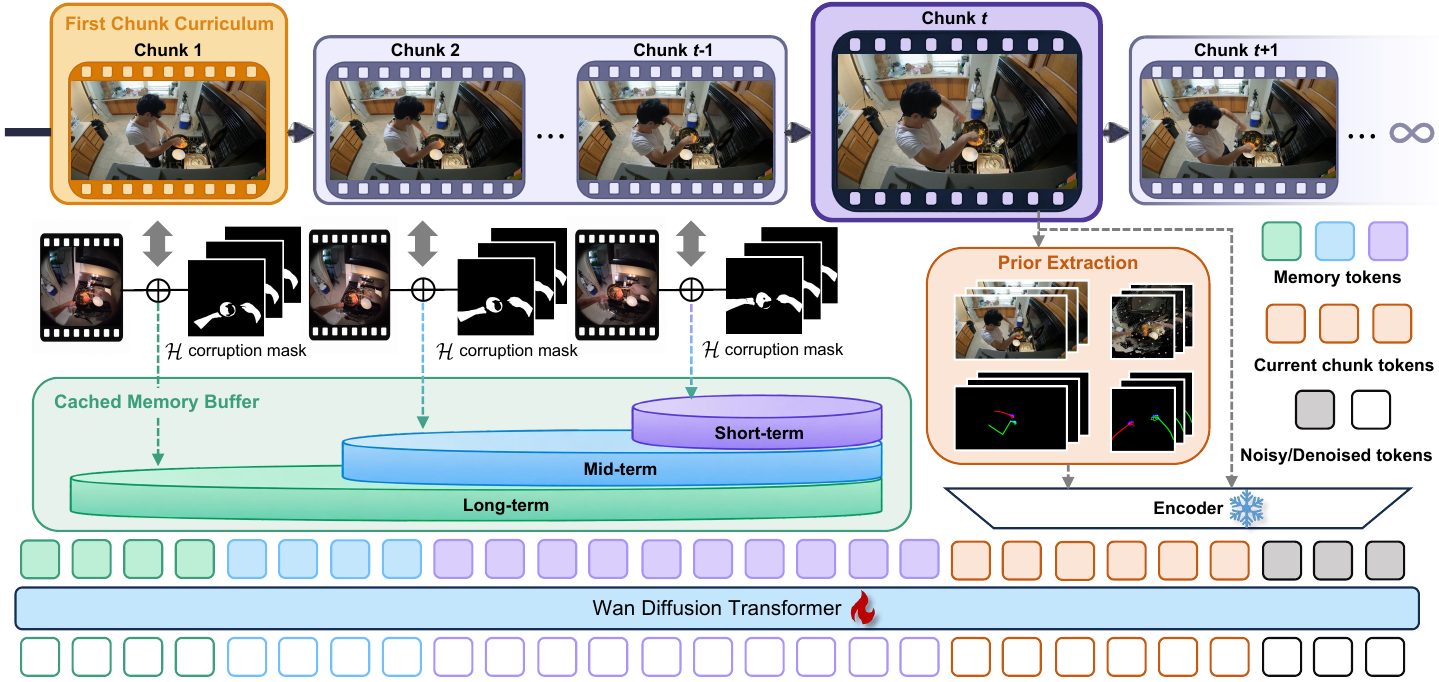}
\caption{Training pipeline for the streaming student model. The $k$-th egocentric chunk is denoised by the Wan DiT~\citep{team2025wan} conditioned on its dual-branch priors and the memory buffer. We apply the proposed region-aware corruption and first-chunk curriculum during training.}
\label{fig:pipeline-helios}
\end{figure}

To enable streaming from our bidirectional teacher, we adapt the framework into a causal, chunk-wise autoregressive student distilled for few-step inference. Following Helios~\citep{yuan2026helios}, this adaptation proceeds in three stages: causal attention across chunks, pyramid token compression, and few-step distillation with DMD~\citep{yin2024onestep} and adversarial training (Appendix~\ref{sec:supp_ar_transition}).
As illustrated in Fig.~\ref{fig:pipeline-helios}, the video stream is processed chunk by chunk; the $k$-th chunk (which denotes the current chunk) is denoised via intra-chunk bidirectional attention conditioned on current-chunk priors (Sec.~\ref{sec:dual_condition}) and a historical memory buffer $\mathcal{M}_k$ which maintains exclusively clean egocentric latents from past chunks and compresses distant context via multi-term memory patchification~\citep{yuan2026helios}.
To address Exo2Ego task-specific challenges in autoregressive training, we introduce two dedicated strategies: \emph{region-aware corruption} to mitigate exposure bias and motion drift in $\mathcal{H}$, and a \emph{first-chunk curriculum} to handle the initial chunk without historical conditioning, improving the student's rollout robustness in $\mathcal{U}$.

\noindent \textbf{Region-aware corruption.}
The exposure bias induced by teacher forcing is highly uneven across regions: $\mathcal{S}$ is continuously anchored through the proposed streaming dynamic ego prior, resilient to drift.
By contrast, the dynamic and complex nature of region $\mathcal{H}$ makes it susceptible to self-occlusions and motion blur, under which the streaming priors can degrade. In such unconstrained cases, synthesis in $\mathcal{H}$ refers to historical context and thus prone to error accumulation.

We inject stochastic perturbations into $\mathcal{H}$ on historical latents $z_t^g \in \mathcal{M}_k$:
\begin{equation}
\widetilde{z}_t^g = z_t^g + b_t m_t^H \odot \epsilon_t,
\label{eq:ar_corruption}
\end{equation}
where $m_t^H$ is the binary $\mathcal{H}$ mask, $b_t \sim \operatorname{Bernoulli}(p)$ acts as a stochastic trigger, and zero-mean Gaussian noise $\epsilon_t$ injects additive perturbation. This mechanism forces the student to ground hand dynamics in current condition priors rather than over-relying on potentially degraded history.

\noindent \textbf{First-chunk curriculum.}
Because region $\mathcal{U}$ lacks direct exocentric observations, subsequent chunks rely entirely on previously synthesized ego frames for consistency. Thus, rollout quality in $\mathcal{U}$ heavily depends on the cold-start synthesis of the first chunk ($k=1, \mathcal{M}_1 = \emptyset$). Any initial geometric artifacts in region $\mathcal{U}$ propagate downstream as erroneous history, triggering degradations.

We therefore introduce a progressive training curriculum on chunk sampling. We initially train exclusively on the first chunk to anchor high-fidelity inpainting from visual priors alone. As training proceeds, its sampling probability decays linearly, shifting focus toward continuation chunks. This design ensures robust visual initialization before exposing the model to autoregressive training.

\noindent \textbf{Few-step distillation.}
We adapt the few-step Distribution Matching Distillation from Helios~\citep{yuan2026helios} to distill our causal teacher $G_\theta$ into a few-step student generator $G_\phi$:
\begin{equation}
\begin{aligned}
\mathcal{L}_{\mathrm{student}} &= \mathcal{L}_{\mathrm{DMD}}(G_\phi; \mu_{\mathrm{real}}, \mu_{\mathrm{fake}}) + \lambda_G \mathcal{L}_{\mathrm{adv}}^G(G_\phi, D_\psi), \\
\mathcal{L}_{\mathrm{critic}} &= \mathcal{L}_{\mathrm{reg}}(D_\psi) + \lambda_D \mathcal{L}_{\mathrm{adv}}^D(G_\phi, D_\psi) + \lambda_R \mathcal{L}_{\mathrm{R1}}(D_\psi),
\end{aligned}
\label{eq:ar_distillation}
\end{equation}
where $\mathcal{L}_{\mathrm{DMD}}$ aligns the generative distribution, and the adversarial objective with discriminator $D_\psi$ sharpens fine details. We refer the readers to the Appendix for more details.

\section{Experiments}
\label{sec:experiments}

\subsection{Implementation details}
\label{sec:exp_setup}

\noindent \textbf{Training.}
We train both our bidirectional teacher and autoregressive student on 4,400 curated clips from Ego-Exo4D~\citep{grauman2024egoexo4d}, with each clip spanning 5 seconds at 24 FPS. Exocentric inputs are set to a resolution of $832\times480$, while generated egocentric outputs are $480\times480$. The bidirectional teacher (Sec.~\ref{sec:method_teacher}) is initialized from Wan2.2-TI2V-5B~\citep{team2025wan}, and the student model (Sec.~\ref{sec:method_student}) is initialized directly from this trained teacher. Detailed data curation procedures and training configurations are provided in the Appendix.

\noindent \textbf{Evaluation data.}
We evaluate quantitatively on Ego-Exo4D~\citep{grauman2024egoexo4d} using both the \textsc{Unseen} (unseen scene during training) and \textsc{New Action} (seen scene with unseen actions) splits as well as H2O~\citep{h2o} (with ground truth calibrated to the FOV of \cite{grauman2024egoexo4d}). We evaluate only qualitatively on Assembly101~\citep{sener2022assembly101}, as its egocentric views provide only monochromatic ground truth within a single indoor scene.

\noindent \textbf{Evaluation metrics.}
We evaluate performance across four groups:
(1) \emph{Image metrics}: $\mathcal{S}$-PSNR, $\mathcal{S}$-MSSIM~\citep{mssim}, $\mathcal{S}$-LPIPS measure fidelity in $\mathcal{S}$, $\mathcal{U}$-CLIP-I measure naturalness and visual plausibility of $\mathcal{U}$ and PCK@0.05~\citep{yang2012articulated} and Motion-Consistency (MC) measures hand position accuracy and motion accuracy in $\mathcal{H}$. Full-image PSNR and SSIM are also provided for reference.
(2) \emph{Video metrics}: FVD evaluate distribution discrepancy and inter-frame temporal continuity. Reference-free VBench~\citep{zhang2025evaluation} dimensions temporal flickering (TF) and motion smoothness (MS) evaluates temporal quality and motion quality.
(3) \emph{Object metrics}: Following the matching pipeline in EgoX~\citep{kang2025egoxegocentricvideogeneration} using SAM~2 and DINOv3, we report center location error and bounding-box IoU.
Additionally, we measure generation efficiency by inference FPS.
(4) \emph{Efficiency}: Inference FPS.

\noindent \textbf{Baselines.}
We compare against Exo2Ego baselines, including EgoX~\citep{kang2025egoxegocentricvideogeneration} and Exo2Ego-V~\citep{liu2024exocentric}, alongside representative camera-control models: TrajCrafter~\citep{yu2025trajectorycrafter}, Wan-VACE~\citep{jiang2025vace}, and Vista4D~\citep{lin2026vista4d}. For a fair comparison, all baselines are fine-tuned on our training dataset. Other related works~\citep{xu2025egoexogen,mahdi2025exo2egosyn,wang2026groundedexo2ego,mahdi2026synchronysequenceexotoegogeneration} are omitted since they are not open-source.

\subsection{Qualitative and quantitative comparison}
\label{sec:exp_comparison}

\subsubsection{Short video}
\label{sec:exp_short}

\begin{figure*}[t]
  \centering
  \includegraphics[width=\textwidth]{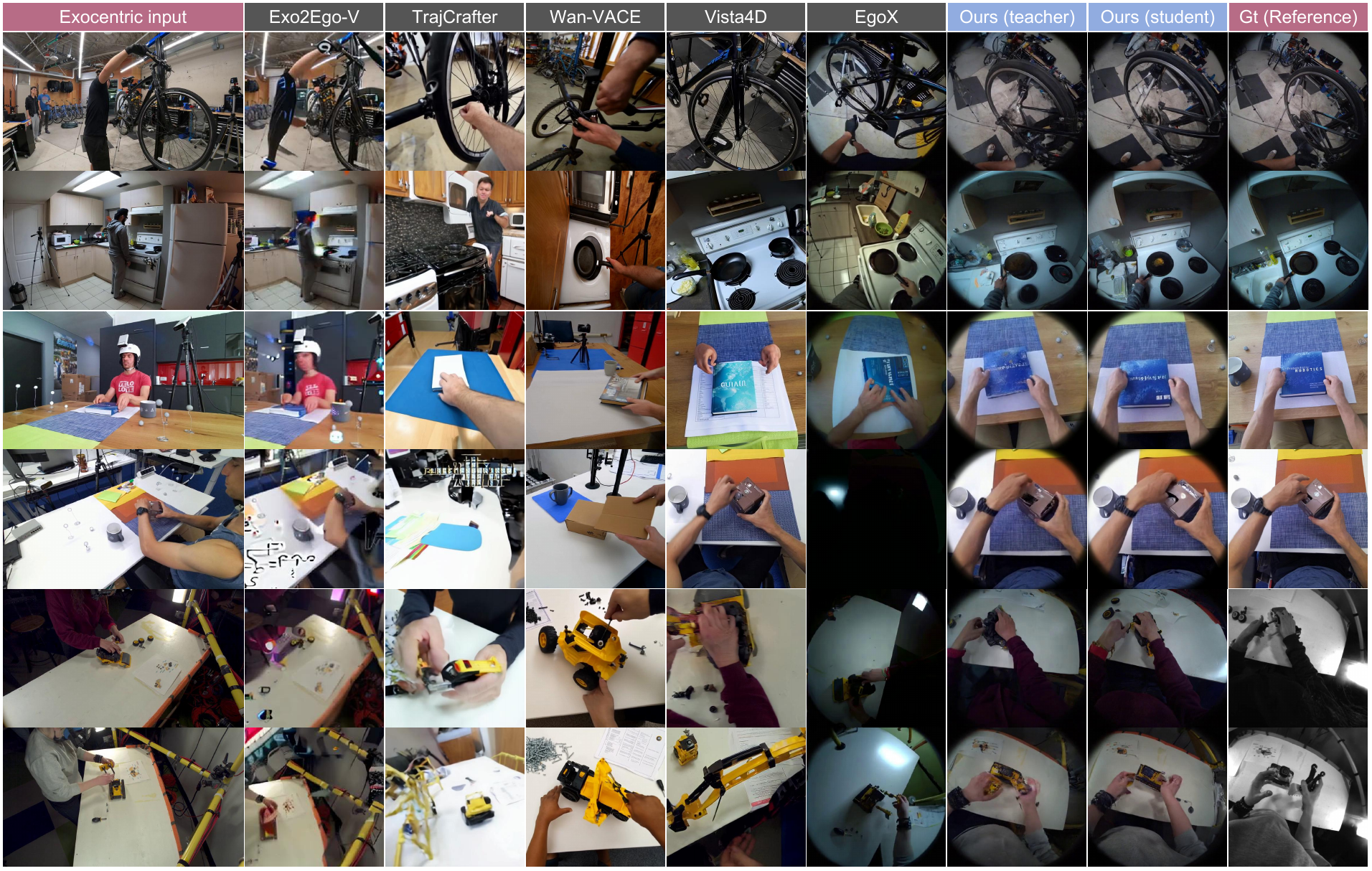}
  \caption{Visual comparison from \cite{grauman2024egoexo4d} (\textsc{Unseen}), \cite{h2o}, and \cite{sener2022assembly101} on synchronized key frames.}
  \label{fig:short_video_comparison}
\end{figure*}

\begin{table}[t]
  \centering
  \caption{Comparison on the \textsc{New Action} and \textsc{Unseen} from Ego-Exo4D~\citep{grauman2024egoexo4d} and H2O~\citep{h2o}. \textbf{Bold}: the best. \underline{Underline}: second best. Ours (est.) runs Ours (long) with ego trajectories estimated from the exocentric video (Sec.~\ref{sec:exp_e2e}).}
  \label{tab:short_video_comparison}
  \setlength{\tabcolsep}{3.6pt}
  \resizebox{\linewidth}{!}{%
  \begin{tabular}{llcccccccccccccc}
    \toprule
    & \multirow{2}{*}{Method} & \multicolumn{8}{c}{Image Metrics} & \multicolumn{2}{c}{Object Metrics} & \multicolumn{3}{c}{Video Metrics} & Efficiency \\
    \cmidrule(lr){3-10}\cmidrule(lr){11-12}\cmidrule(lr){13-15}\cmidrule(lr){16-16}
    & & $\mathcal{S}$-PSNR $\uparrow$ & $\mathcal{S}$-MSSIM $\uparrow$ & $\mathcal{S}$-LPIPS $\downarrow$ & $\mathcal{U}$-CLIP-I $\uparrow$ & PCK@0.05 $\uparrow$ & MC $\uparrow$ & PSNR $\uparrow$ & SSIM $\uparrow$ & Loc Err $\downarrow$ & IoU $\uparrow$ & FVD $\downarrow$ & TF $\uparrow$ & MS $\uparrow$ & FPS $\uparrow$ \\
    \midrule
    \multirow{9}{*}{\rotatebox{90}{\shortstack{\textsc{New Action}\\{\fontsize{7}{8}\selectfont (new actions, seen envs.)}}}}
    & Exo2Ego-V & 13.84 & 0.406 & 0.558 & 0.804 & 15.23\% & 17.47\% & 14.37 & 0.393 & 0.315 & 0.098 & 605.90 & 0.963 & 0.965 & \underline{3.92} \\
    & TrajCrafter & 10.91 & 0.347 & 0.617 & 0.783 & 3.60\% & 16.79\% & 12.55 & 0.349 & 0.232 & 0.116 & 568.92 & 0.960 & 0.980 & 0.30 \\
    & Wan-VACE & 10.42 & 0.383 & 0.624 & 0.817 & 5.83\% & 18.37\% & 13.23 & 0.430 & 0.230 & 0.120 & 505.89 & \textbf{0.989} & \textbf{0.994} & 0.96 \\
    & Vista4D & 12.07 & 0.412 & 0.465 & 0.801 & 11.58\% & 15.92\% & 13.42 & 0.435 & 0.135 & 0.249 & 353.83 & 0.983 & 0.992 & 0.07 \\
    & EgoX & 15.75 & 0.489 & 0.580 & 0.888 & 25.91\% & 34.84\% & 16.06 & 0.561 & 0.130 & 0.357 & 184.24 & 0.977 & 0.990 & 0.08 \\
    & Ours (teacher) & \textbf{18.35} & \textbf{0.603} & 0.372 & \textbf{0.907} & \textbf{49.86\%} & \textbf{56.26\%} & \textbf{17.51} & \textbf{0.603} & \textbf{0.076} & \textbf{0.478} & \textbf{111.97} & 0.983 & \textbf{0.994} & 0.97 \\
    & Ours (student) & 18.12 & \underline{0.599} & \textbf{0.366} & 0.891 & \underline{47.61\%} & 53.12\% & 17.45 & \underline{0.593} & \underline{0.099} & \underline{0.462} & \underline{128.52} & \underline{0.984} & \underline{0.993} & \textbf{24.3} \\
    & Ours (long) & \underline{18.16} & 0.588 & \underline{0.365} & \underline{0.893} & 47.59\% & \underline{53.48\%} & \underline{17.47} & 0.588 & 0.118 & 0.447 & 135.62 & 0.981 & 0.991 & \textbf{24.3} \\
    & Ours (est.) &16.11 & 0.469 & 0.491 & 0.865 & 30.98\% & 36.47\% & 15.26 & 0.450 & 0.164 & 0.308 & 343.55 & 0.973 & 0.982 & \textbf{24.3} \\
    \midrule
    \multirow{9}{*}{\rotatebox{90}{\shortstack{\textsc{Unseen}\\{\fontsize{7}{8}\selectfont (new actions, new envs.)}}}}
    & Exo2Ego-V & 13.06 & 0.425 & 0.581 & 0.691 & 11.57\% & 13.26\% & 12.94 & 0.385 & 0.461 & 0.003 & 1237.55 & 0.973 & 0.978 & \underline{3.92} \\
    & TrajCrafter & 10.73 & 0.295 & 0.607 & 0.782 & 2.75\% & 12.65\% & 12.31 & 0.306 & 0.386 & 0.041 & 818.63 & 0.966 & 0.984 & 0.30 \\
    & Wan-VACE & 10.04 & 0.352 & 0.649 & 0.825 & 5.41\% & 17.22\% & 12.21 & 0.378 & 0.412 & 0.038 & 1023.96 & \textbf{0.995} & \textbf{0.996} & 0.96 \\
    & Vista4D & 11.58 & 0.401 & 0.576 & 0.840 & 9.37\% & 12.08\% & 12.64 & 0.435 & 0.144 & 0.230 & 783.54 & \underline{0.990} & 0.994 & 0.07 \\
    & EgoX & 13.36 & 0.406 & 0.577 & 0.879 & 11.07\% & 15.41\% & 14.41 & 0.456 & 0.311 & 0.092 & 438.79 & 0.981 & 0.992 & 0.08 \\
    & Ours (teacher) & \textbf{18.18} & \underline{0.576} & \underline{0.405} & \textbf{0.901} & \textbf{40.24\%} & \textbf{43.76\%} & \textbf{17.19} & \underline{0.544} & \textbf{0.106} & \textbf{0.456} & \underline{138.91} & 0.980 & \underline{0.995} & 0.97 \\
    & Ours (student) & \underline{18.04} & \textbf{0.583} & 0.423 & \underline{0.894} & \underline{37.49\%} & 41.13\% & \underline{17.08} & \textbf{0.548} & \underline{0.113} & \underline{0.436} & \textbf{136.48} & 0.986 & 0.990 & \textbf{24.3} \\
    & Ours (long) & 17.88 & 0.569 & \textbf{0.394} & 0.890 & 37.22\% & \underline{41.70\%} & 16.84 & 0.537 & 0.119 & 0.427 & 144.33 & 0.988 & 0.994 & \textbf{24.3} \\
    & Ours (est.) &15.28 & 0.467 & 0.517 & 0.848 & 23.76\% & 31.32\% & 14.70 & 0.443 & 0.183 & 0.290 & 368.42 & 0.972 & 0.979 & \textbf{24.3} \\
    \midrule
    \multirow{9}{*}{\rotatebox{90}{\shortstack{\cite{h2o}\\{\fontsize{7}{8}\selectfont (in-the-wild.)}}}}
    & Exo2Ego-V & 12.25 & 0.402 & 0.493 & 0.675 & 12.39\% & 15.14\% & 12.34 & 0.411 & 0.445 & 0.025 & 882.92 & 0.965 & 0.972 & \underline{3.92} \\
    & TrajCrafter & 12.98 & 0.379 & 0.465 & 0.774 & 20.60\% & 62.96\% & 13.15 & 0.391 & 0.362 & 0.163 & 724.07 & 0.961 & 0.980 & 0.30 \\
    & Wan-VACE & 14.78 & 0.348 & 0.322 & 0.804 & 27.35\% & 68.66\% & 14.65 & 0.376 & 0.310 & 0.183 & 662.87 & \textbf{0.995} & \textbf{0.996} & 0.96 \\
    & Vista4D & 18.39 & 0.644 & 0.181 & 0.916 & 68.57\% & 52.99\% & 18.31 & 0.640 & 0.072 & 0.741 & 161.45 & 0.990 & \underline{0.994} & 0.07 \\
    & EgoX & 14.23 & 0.465 & 0.377 & 0.710 & 35.10\% & 40.28\% & 15.00 & 0.468 & 0.205 & 0.513 & 479.52 & 0.911 & 0.929 & 0.08 \\
    & Ours (teacher) & \textbf{23.59} & \textbf{0.724} & \textbf{0.106} & \textbf{0.931} & \textbf{89.22\%} & \textbf{95.78\%} & \textbf{21.35} & \textbf{0.688} & \textbf{0.041} & \textbf{0.876} & \textbf{89.94} & \underline{0.993} & \underline{0.994} & 0.97 \\
    & Ours (student) & \underline{23.42} & \underline{0.713} & \underline{0.118} & 0.926 & \underline{84.40\%} & \underline{93.14\%} & 20.92 & \underline{0.680} & \underline{0.045} & \underline{0.859} & \underline{93.47} & \underline{0.993} & \underline{0.994} & \textbf{24.3} \\
    & Ours (long) & 23.21 & \underline{0.713} & \underline{0.118} & \underline{0.929} & 81.80\% & 89.71\% & \underline{21.05} & \underline{0.680} & 0.056 & 0.814 & 94.17 & \underline{0.993} & \underline{0.994} & \textbf{24.3} \\
    & Ours (est.) &20.28 & 0.668 & 0.179 & 0.871 & 63.50\% & 71.26\% & 18.56 & 0.615 & 0.115 & 0.744 & 174.01 & 0.975 & 0.984 & \textbf{24.3} \\
    \bottomrule
  \end{tabular}}
\end{table}

For a fair evaluation, Ours (teacher), Ours (student), and all baseline methods are compared on short video clips truncated to identical temporal durations in Table.~\ref{tab:short_video_comparison}.
In addition, Ours (long) reports scores computed from our student model running for 30-second continuous rollouts initialized from the same starting timestamps across the test splits.

As shown in Table~\ref{tab:short_video_comparison} and Figure~\ref{fig:short_video_comparison}, StreamEgo achieves superior performance across quantitative metrics and visual comparisons.:
(1) Across image- and object-level metrics, our framework aligns closely with the task objectives of Exo2Ego: establishing high visual fidelity in region $\mathcal{S}$, maintaining plausible naturalness and diversity region $\mathcal{U}$, and ensuring precise hand articulation and motion consistency in region $\mathcal{H}$.
(2) As evidenced by both the metrics (PCK@0.05 and MC in Table.~\ref{tab:short_video_comparison}) and visual results, StreamEgo substantially outperforms baseline methods in hand-motion accuracy, thanks to the proposed streaming dual-branch prior providing appearance and structural guidance for region $\mathcal{H}$.
(3) Camera control models~\citep{yu2025trajectorycrafter,jiang2025vace,lin2026vista4d} largely fail due to their underlying assumption of modest viewpoint changes. Besides, existing Exo2Ego methods like \cite{kang2025egoxegocentricvideogeneration} suffer from viewpoint rotational drifts and degrade severely in out-of-distribution cases. In contrast, our method demonstrates markedly superior cross-domain generalization on in-the-wild datasets~\citep{h2o,sener2022assembly101}, validating the effectiveness of our region-aware curriculum in coordinating multi-region optimization.
(4) In terms of efficiency, our distilled student model achieves real-time streaming speeds while maintaining competitive quality against the teacher model, underscoring its potential for practical applications.

\subsubsection{Long video}
\label{sec:exp_long}

\begin{figure*}[t]
  \centering
  \includegraphics[width=\textwidth]{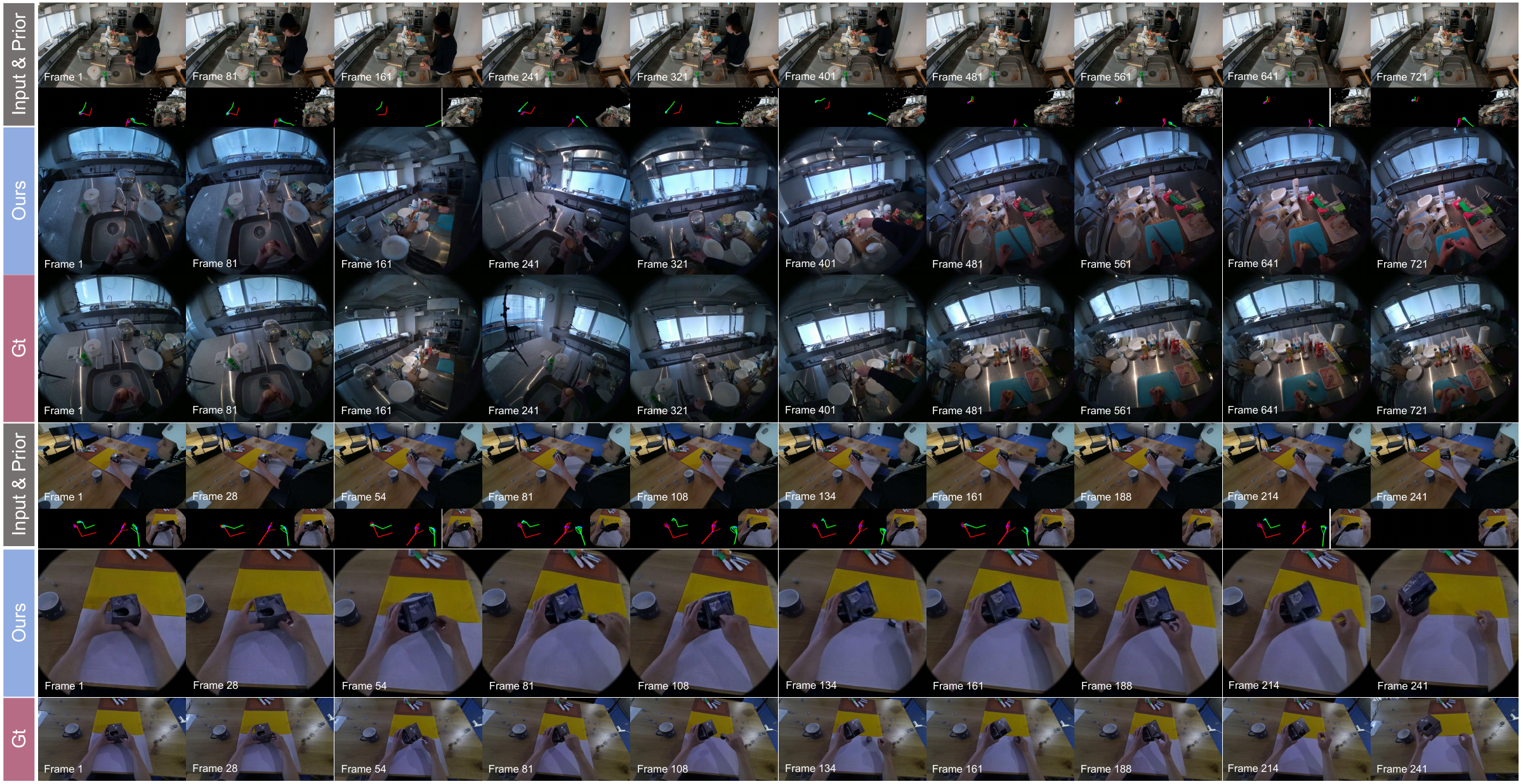}
  \caption{Streaming long-video generation with the autoregressive student on \cite{grauman2024egoexo4d} (\textsc{Unseen}) and \cite{h2o}.}
  \label{fig:long_video_results}
\end{figure*}

As shown in Fig.~\ref{fig:long_video_results}, our student operates as a causal streaming model, taking input chunk-by-chunk and generating the corresponding egocentric chunk with only a single-chunk warm-up latency (\textit{i.e.}, emitting the first egocentric chunk as the second exocentric chunk arrives. With chunk size $33$ frames, this corresponds to roughly $1.375$ seconds.).
The generated long videos demonstrate sustained visual quality, smooth temporal coherence, and minimal flicker, remaining highly comparable to the teacher model in both viewpoint consistency and hand-motion accuracy (Table~\ref{tab:short_video_comparison}: Ours (Long)).
Moreover, the model generalizes robustly to in-the-wild recordings such as H2O~\citep{h2o}.
Furthermore, generation quality remains stable throughout the rollouts.
This resilience stems from the strong initial transfer inherited from the bidirectional teacher, combined with continuous, chunk-wise pixel-aligned anchors in region $\mathcal{S}$, while our first-chunk curriculum and region-aware corruption effectively mitigate exposure bias across regions $\mathcal{U}$ and $\mathcal{H}$.

\subsection{Ablation studies}
\label{sec:exp_ablation}

As summarized in Table~\ref{tab:ablation}, we evaluate key components against the full teacher and student models. For the teacher, we isolate the loss curriculum (\textit{w/o} HSU) and condition priors: removing the streaming human pose prior (\textit{w/o pose}), replacing our body-guided pose prior with isolated hand poses~\citep{park2026egoworld} (Ew pose) or SAM-based hand-and-object predictions~\citep{xu2025egoexogen} (HOI pose), and removing the streaming dynamic prior (\textit{w/o} prior), substituting our streaming dynamic prior with EgoX's offline ego prior (E prior) lacking foreground segmentation. For the student, we ablate region-aware history corruption (\textit{w/o} H corruption) and the first-chunk curriculum (\textit{w/o} first chunk) to verify their roles in stable autoregressive rollout.

\begin{figure*}[t]
  \centering
  \includegraphics[width=\linewidth]{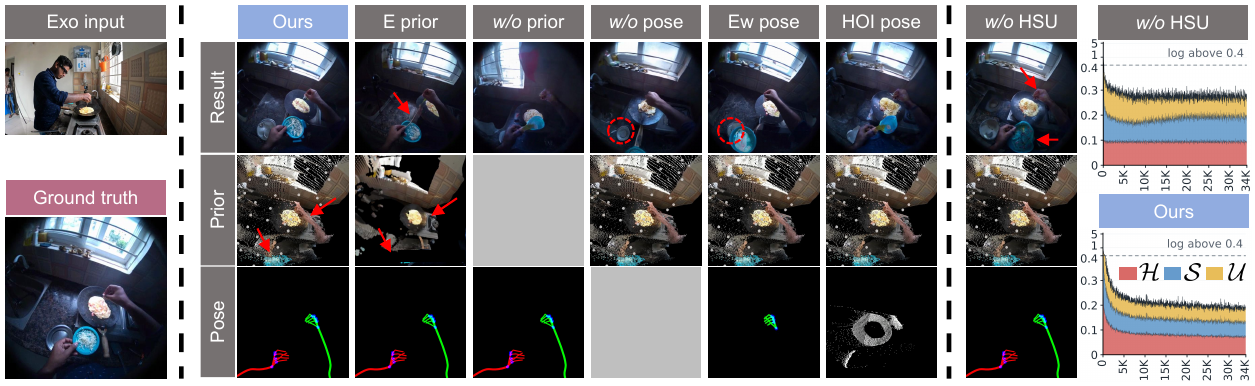}
  \caption{Ablation comparison on \cite{grauman2024egoexo4d}. \textbf{Left:} exocentric input and its egocentric ground truth. \textbf{Middle:} ablations results for the streaming dual-branch prior. \textbf{Right:} ablation results and training log for the region-aware curriculum. Red circles and arrows mark failure regions.}
  \label{fig:ablation_comparison}
\end{figure*}

\begin{table}[t]
  \centering
  \caption{Ablation study on Ego-Exo4D~\citep{grauman2024egoexo4d} \textsc{Unseen}. Rows marked $\dagger$ evaluate region-aware loss weight scheduling on \textsc{New Action}. Teacher variants isolate conditioning and loss scheduling; student variants isolate autoregressive training strategies. \textbf{Bold: The best}.}
  \label{tab:ablation}
  \renewcommand{\arraystretch}{0.9}
  \setlength{\tabcolsep}{5pt}
  \resizebox{\linewidth}{!}{%
  \begin{tabular}{lccccccccccc}
    \toprule
    \multirow{2}{*}{Variant} & \multicolumn{6}{c}{Image Metrics} & \multicolumn{2}{c}{Object Metrics} & \multicolumn{3}{c}{Video Metrics} \\
    \cmidrule(lr){2-7}\cmidrule(lr){8-9}\cmidrule(lr){10-12}
    & $\mathcal{S}$-PSNR $\uparrow$ & $\mathcal{S}$-SSIM $\uparrow$ & $\mathcal{S}$-LPIPS $\downarrow$ & $\mathcal{U}$-CLIP-I $\uparrow$ & PCK@0.05 $\uparrow$ & MC $\uparrow$ & Loc Err $\downarrow$ & IoU $\uparrow$ & FVD $\downarrow$ & TF $\uparrow$ & MS $\uparrow$ \\
    \midrule
    Full teacher & \textbf{18.18} & \textbf{0.576} & \textbf{0.405} & \textbf{0.901} & \textbf{40.24\%} & \textbf{43.76\%} & \textbf{0.106} & \textbf{0.456} & \textbf{138.91} & \textbf0.980 & \textbf{0.995} \\
    w/o pose & 18.10 & 0.564 & 0.430 & 0.893 & 16.75\% & 19.27\% & 0.110 & 0.448 & 284.59 & 0.978 & 0.991 \\
    Ew pose & 18.14 & 0.572 & 0.422 & 0.887 & 18.03\% & 29.64\% & 0.115 & 0.452 & 289.04 & 0.980 & 0.988 \\
    HOI pose & 17.61 & 0.558 & 0.444 & 0.880 & 17.59\% & 24.80\% & 0.172 & 0.341 & 275.50 & 0.968 & 0.985 \\
    w/o prior & 13.57 & 0.396 & 0.588 & 0.714 & 15.71\% & 18.35\% & 0.255 & 0.126 & 645.42 & 0.959 & 0.962 \\
    E prior & 18.05 & 0.546 & 0.411 & 0.890 & 36.16\% & 40.38\% & 0.198 & 0.366 & 144.45 & 0.980 & 0.991 \\
    w/o HSU & 16.88 & 0.481 & 0.514 & 0.864 & 29.22\% & 31.37\% & 0.222 & 0.139 & 301.76 & 0.977 & 0.989 \\
    \midrule
    Full student & \textbf{17.88} & \textbf{0.569} & \textbf{0.394} & \textbf{0.890} & \textbf{37.22\%} & \textbf{41.70\%} & \textbf{0.119} & \textbf{0.427} & \textbf{144.33} & \textbf{0.988} & \textbf{0.994} \\
    w/o H corruption & 17.84 & 0.560 & 0.412 & 0.875 & 31.27\% & 35.84\% & 0.142 & 0.400 & 203.48 & 0.984 & 0.992 \\
    w/o first-chunk & 17.32 & 0.557 & 0.395 & 0.837 & 36.90\% & 38.73\% & 0.294 & 0.273 & 397.10 & 0.981 & 0.987 \\
    \midrule
    Full teacher$^\dagger$ & \textbf{18.35} & \textbf{0.603} & \textbf{0.372} & \textbf{0.907} & \textbf{49.86\%} & \textbf{56.26\%} & \textbf{0.076} & \textbf{0.478} & \textbf{111.97} & \textbf{0.983} & \textbf{0.994} \\
    w/o HSU$^\dagger$ & 17.64 & 0.531 & 0.465 & 0.902 & 35.49\% & 42.70\% & 0.150 & 0.291 & 144.97 & 0.976 & 0.992 \\
    \bottomrule
  \end{tabular}}
\end{table}

\noindent \textbf{Human pose prior.}
Removing the streaming human pose prior leads to position inaccuracy and motion degradation in $\mathcal{H}$ (PCK@0.05 and MC in Table.~\ref{tab:ablation}; missing left hand in Fig.~\ref{fig:ablation_comparison}: \textit{w/o} pose), which can not be recovered by adopting isolated hand pose estimation~\citep{park2026egoworld} with inaccurate projected hand positions (Fig.~\ref{fig:ablation_comparison}: Ew pose). Substituting our prior with HOI mask predictions~\citep{xu2025egoexogen} produces blurred manipulations since it can not define the structure of $\mathcal{H}$. %

\noindent \textbf{Dynamic ego prior.}
Removing the dynamic ego prior eliminates pixel-aligned appearance anchors, causing severe visual degradation in $\mathcal{S}$ and noticeable viewpoint drift. Replacing it with EgoX's prior removes appearance guidance in $\mathcal{H}$, leading to degraded hand quality (Fig.~\ref{fig:ablation_comparison}: E prior).

\noindent \textbf{Loss scheduling.}
Replacing our curriculum with fixed uniform weights leads to inaccurate motion in $\mathcal{H}$, alongside degradations in $\mathcal{S}$ and $\mathcal{U}$ (Fig.~\ref{fig:ablation_comparison}: blurry hands and distorted structures). The loss curves (Fig.~\ref{fig:ablation_comparison}, \textbf{right}) reveal that \textit{w/o} HSU hinders convergence in $\mathcal{H}$ and $\mathcal{S}$ (\textit{e.g.}, lower $\mathcal{S}$-PSNR and MC in Table~\ref{tab:ablation}) while inducing premature overfitting in $\mathcal{U}$. In contrast, our curriculum stabilizes training and enhances generalization. From another perspective, this is also evidenced by Table~\ref{tab:short_video_comparison}: our method achieves comparable $\mathcal{S}$-PSNR, $\mathcal{S}$-MSSIM, and $\mathcal{S}$-LPIPS across \textsc{New Action} and \textsc{Unseen}, whereas full-image PSNR and SSIM drop on \textsc{Unseen}. This indicates lower pixel-level agreement in the unobservable region $\mathcal{U}$, while $\mathcal{U}$-CLIP-I remains consistent across both splits, confirming that our model performs plausible, semantics-preserving inpainting in $\mathcal{U}$ rather than memorizing training scenes. Additionally, the ablation in Table~\ref{tab:ablation} (\textit{w/o} HSU$^\dagger$) further demonstrates that this generalization capability stems from our region-aware curriculum: without it, performance exhibits a pronounced gap between \textsc{New Action} and \textsc{Unseen}, reflecting overfitting patterns to seen environments.

\begin{wrapfigure}{!b}{0.5\textwidth}
\centering
\vspace{-15pt}
\includegraphics[width=\linewidth]{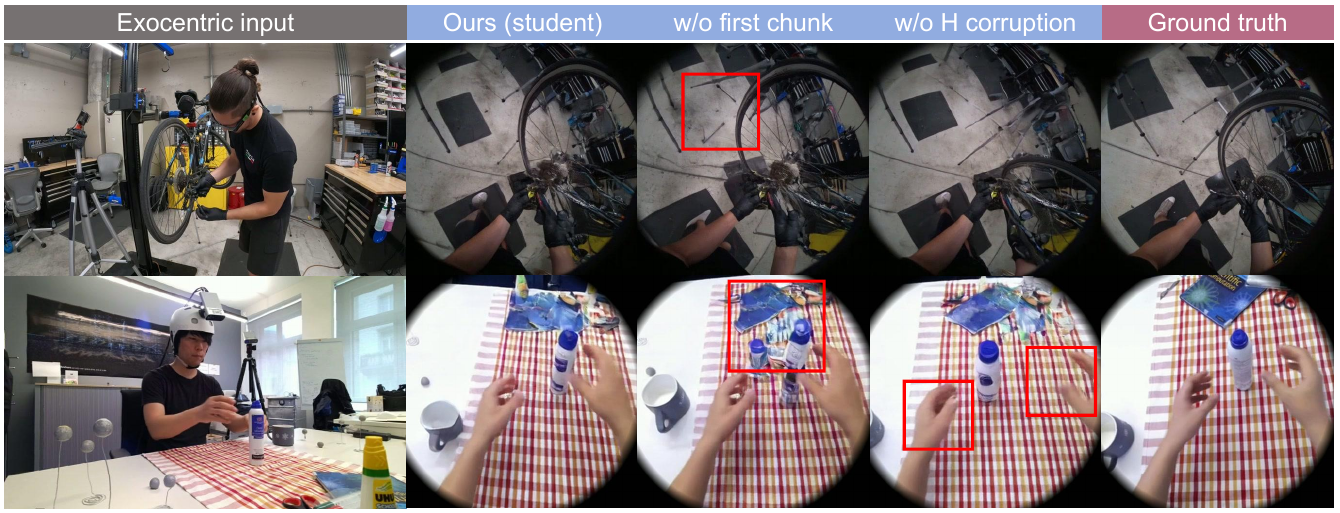}
\caption{Ablation comparisons for region-aware corruption and first-chunk curriculum.}
\vspace{-15pt}
\label{fig:ablation-student}
\end{wrapfigure}

\noindent \textbf{Region-aware corruption.}
Disabling the region-aware corruption leaves $\mathcal{H}$ vulnerable to exposure bias between teacher-forced training and free-running rollout. As is shown in Fig.~\ref{fig:ablation-student}, the hand position drifts, leading to degradation to $\mathcal{H}$-region image metrics in Table.~\ref{tab:ablation}.

\noindent \textbf{First-chunk curriculum.}
Removing the first-chunk curriculum degrades yields blur and distortion in $\mathcal{U}$ (Fig.~\ref{fig:ablation-student}: distorted book and bottles; $\mathcal{U}$-CLIP-I in Table.~\ref{fig:ablation_comparison}). This stems from two interrelated factors: synthesizing the initial chunk is inherently harder without historical reference, and any resulting artifacts may propagate, compounding errors across subsequent chunks.

\subsection{End-to-end streaming with estimated ego trajectories}
\label{sec:exp_e2e}

Like all prior Exo2Ego work~\citep{kang2025egoxegocentricvideogeneration,xu2025egoexogen,mahdi2025exo2egosyn,park2026egoworld,mahdi2026synchronysequenceexotoegogeneration,wang2026groundedexo2ego}, we take the target egocentric camera trajectory as a given input (As is stated in Sec.~\ref{sec:method}), and the experiments above use the trajectories provided by the datasets~\citep{h2o,grauman2024egoexo4d,sener2022assembly101}. These trajectories come from the tracked first-person capture of the datasets, which is not available when only an exocentric camera observes the scene. A practical streaming system therefore needs to obtain the egocentric trajectory online from the exocentric video, at the same pace as the video arrives. To test StreamEgo as such a complete system, we implement and evaluate a fully end-to-end pipeline that estimates the egocentric trajectory from the exocentric video and takes no ground-truth trajectory as input.

\noindent \textbf{Head tracker.} We replace the dataset trajectories with head poses estimated by a causal tracker built from open-source estimators. Multi-HMR~\citep{baradel2024multihmr} estimates the 3D head position on every second frame, MoGe-2~\citep{wang2025moge2} aligns its metric depth, SemiUHPE~\citep{zhou2025semiuhpe} estimates the head orientation, and a causal One Euro filter~\citep{casiez2012oneeuro} smooths the pose trajectory. A VLM identifies the wearer once at the start (More details are in Appendix~\ref{sec:supp_tracker}). The tracker takes 0.91\,s per 1.375-s chunk (33 frames at $832\times480$ resolution) on an A100 GPU, which is well within the chunk duration. The end-to-end pipelined throughput therefore stays at 24.3 FPS. See Table~\ref{tab:supp_tracker_latency} for a latency breakdown.

\noindent \textbf{Quality.} The estimated poses deviate from the dataset trajectories by 52.8\,cm and 22.3$^\circ$ on \textsc{New Action}, by 54.5\,cm and 22.7$^\circ$ on \textsc{Unseen}, and by 6.3\,cm and 10.9$^\circ$ on H2O. On Ego-Exo4D, the position error lies mostly along the viewing ray, as the wearers stand about 3\,m from the camera, compared with about 1\,m on H2O. We run our student with the estimated trajectories under the same 30-second protocol as Ours (long) and report the results as Ours (est.) in Table~\ref{tab:short_video_comparison}. Although the estimated poses are less accurate than the ground-truth trajectories, StreamEgo remains competitive with the baselines in Table~\ref{tab:short_video_comparison}, which all use the ground-truth trajectories. In practical deployments equipped with dedicated hardware trackers (\textit{e.g.}, on-device SLAM or inertial sensors), pose accuracy and thus generation quality can be further improved.

\section{Conclusion}
\label{sec:conclusion}

In this work, we presented StreamEgo, a streaming framework enabling unbounded and generalizable exocentric-to-egocentric video generation. To overcome the degraded interactions, overfitting, and fixed-length latency bottlenecks of existing approaches, we proposed three core techniques: streaming dual-branch priors for structurally guided hand-object synthesis, a region-aware curriculum that stabilizes training across functional canvas domains, and a chunk-wise autoregressive adaptation with few-step distillation for causal rollout, together with a region-aware corruption mechanism and a first-chunk curriculum for enhancing quality. Achieving generalizable streaming generation at 24.3 FPS on a single H100 GPU, StreamEgo provides a practical pathway to repurpose vast third-person demonstration recordings into first-person trajectories for embodied robotics.

\bibliography{references}
\bibliographystyle{authoryear}

\clearpage
\appendix
\section{Data Curation}
\label{sec:supp_data_cleaning}

\subsection{Ego-Exo4D training corpus}
\label{sec:supp_data_egoexo}

Ego-Exo4D~\citep{grauman2024egoexo4d} supplies synchronized calibrated exocentric
(GoPro) and first-person (Aria) RGB streams and is the sole source of our
training data. We extract candidate windows of five seconds at $24$\,fps
($121$ frames, endpoints inclusive) on a one-second stride from
manipulation-oriented activities, \textit{e.g.}\ cooking, bicycle repair,
medical procedures, and musical performance, rotating the Aria stream to its
upright orientation before export.

Selection proceeds from inexpensive metadata checks to visual checks. We
discard takes with missing calibration or trajectory data, and retain windows
whose mean exo-to-ego distance is at most $3$\,m, whose ego camera lies inside
the exocentric frustum in at least $85\%$ of frames, and whose ego trajectory
is sufficiently stable (quality score $\ge 0.7$, angular-velocity standard
deviation $\le 0.25$\,rad/s, mean linear velocity $\le 0.4$\,m/s). The visual
screen then requires an actor that is large and central (person height
$\ge 45\%$ of the image in $\ge 70\%$ of sampled frames), at least one visible
hand in $80\%$ of sampled frames, adequate Laplacian sharpness in both views,
and bounded optical-flow magnitude. Surviving clips are ranked by quality
score; non-maximum suppression within each take and exocentric camera enforces
a $1.5$\,s minimum separation, and per-take caps and activity quotas further
suppress near-duplicates while broadening activity coverage. A final
captioning pass records the action and flags poor subject orientation, hidden
hands, and severe exocentric occlusion. The resulting corpus contains
$4{,}400$ training clips and a $100$-clip validation set.

\subsection{Evaluation splits}
\label{sec:supp_data_splits}

Exo2Ego models~\citep{kang2025egoxegocentricvideogeneration,mahdi2025exo2egosyn,wang2026groundedexo2ego,mahdi2026synchronysequenceexotoegogeneration} must generalize rather than
memorize their paired training data. Our evaluation is therefore designed to
be \emph{hard by construction}: from the same source inventory we construct two
$500$-clip splits that are disjoint from training along the two axes that a
paired dataset makes easiest to exploit.

\noindent\textbf{\textsc{New Action}.} These clips depict action categories
\emph{absent} from the training corpus, but are recorded in environments
(scenes, rooms, and capture setups) that the model has seen. They isolate
whether the model has learned to translate a manipulation from the exocentric
view, as opposed to recalling how a familiar action looked in a familiar room.
Because region $\mathcal{U}$ in these clips can often be completed from memorized
scene context, this split principally stresses fidelity in $\mathcal{H}$.

\noindent\textbf{\textsc{Unseen}.} These clips are recorded in environments
that never occur in training, and their actions are likewise held out. No part
of the egocentric canvas, including the currently unobserved region
$\mathcal{U}$, can be recovered by recalling a training scene, so the model must
infer $\mathcal{U}$ from the exocentric evidence alone. This split therefore
penalizes benchmark memorization.

Both splits are constructed at the level of capture takes, so no clip shares a
take, scene, or subject with the training set.

\subsection{H2O: an in-the-wild evaluation set}
\label{sec:supp_data_h2o}

Quantitative evaluation on Ego-Exo4D~\citep{grauman2024egoexo4d}, even on our held-out splits, remains
within the distribution of a single capture protocol. We therefore additionally
evaluate on H2O~\citep{h2o}, an in-the-wild dataset collected with different
hardware, framing, and subject population. To keep the evaluation broad rather
than convenient, we deliberately maximize coverage along three axes rather than
selecting a small homogeneous subset: we retain the widest available range of
exocentric camera viewpoints, participant identities, and indoor scene types.

Concretely, the screening of H2O proceeds as follows. We require synchronized
exocentric and egocentric streams at a native resolution of $1280\times720$
with per-camera intrinsics and a complete egocentric pose trajectory covering
at least $90\%$ of the sequence, discarding any sequence for which a camera
stream or its calibration is incomplete. We then apply a visual screen
requiring adequate sharpness in both views and non-trivial inter-frame motion,
which removes static or over-blurred windows, followed by the same
temporal-NMS and diversity machinery used for Ego-Exo4D, with per-sequence caps
on the number of clips drawn from any single camera or any single sequence.
Crucially, the caps are set high enough that all four exocentric viewpoints and
all six tabletop scene configurations remain represented in the final set, and
the retained clips span every subject group and dozens of distinct action
labels. We deliberately avoid a per-action cap so that rarer manipulations are
not truncated away, which yields a many-action evaluation rather than a
convenient few-class one.

Because H2O is not natively compatible with our setting, we apply two
calibration steps. First, its original $30$\,fps streams are temporally
resampled to the $24$\,fps rate at which both our model and Ego-Exo4D operate,
so that a chunk spans the same duration and frame count across datasets.
Second, and more fundamentally, H2O's egocentric ground truth is captured with
a rectilinear camera that shares the exocentric sensor's geometry and a
horizontal field of view of roughly $90^\circ$, whereas Ego-Exo4D's ego view is
a wide-angle fisheye. Because our ego canvas, our region decomposition, and our
metrics are all defined with respect to the ego intrinsics, feeding H2O's
original ego footage at its native calibration would change what $\mathcal{H}$,
$\mathcal{S}$, and $\mathcal{U}$ mean and make the comparison to the Ego-Exo4D
numbers meaningless. We therefore reproject H2O's egocentric ground truth onto
a virtual Aria fisheye camera built from the calibrated intrinsics of the
Ego-Exo4D ego sensor at $480\times480$, so that the ego trajectory occupies the
same portion of the canvas and the same region masks remain valid. The
reprojection is a geometric resampling only: no colour correction is applied,
and the circular aperture of the virtual sensor is filled with the
corresponding black corners.

With these two steps, we yield a diverse dataset from H2O~\citep{h2o} with $850$ clips, which functions as a genuine cross-dataset test: neither
its exocentric viewpoints, nor its participants, nor its camera model, nor its
scene layouts match the data on which the model was trained.

\section{Additional Results}
\label{sec:supp_more_results}

\subsection{Qualitative results}

\begin{figure*}[t]
  \centering
  \includegraphics[width=\textwidth]{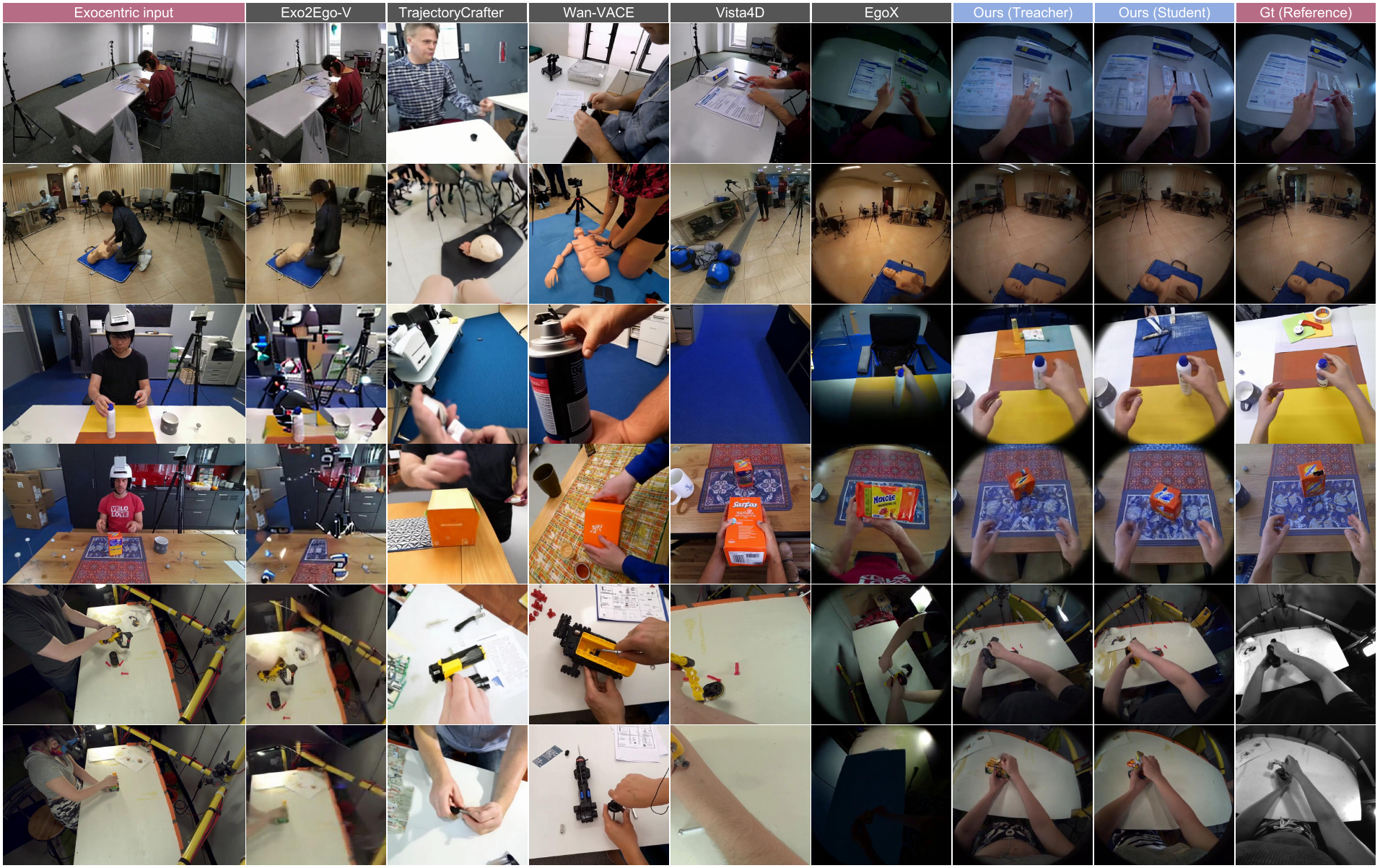}
  \caption{Additional qualitative comparison on synchronized key frames from
  Ego-Exo4D (\textsc{Unseen})~\citep{grauman2024egoexo4d}, H2O~\citep{h2o}, and
  Assembly101~\citep{sener2022assembly101}.}
  \label{supp_fig:short_video_comparison}
\end{figure*}

Fig.~\ref{supp_fig:short_video_comparison} extends the comparison of
Fig.~\ref{fig:short_video_comparison} to additional clips. The trends are
consistent across datasets and splits. Camera-control models
\citep{yu2025trajectorycrafter,jiang2025vace,lin2026vista4d} fails because they work under the assumption of modest viewpoint changes.
Exo2Ego baselines~\citep{liu2024exocentric,kang2025egoxegocentricvideogeneration} recover $\mathcal{S}$ reasonably well but degrade sharply in
$\mathcal{H}$ under self-occlusion, and on \textsc{Unseen} and \cite{h2o,sener2022assembly101} they exhibit the
viewpoint drift or even degrade into darkness. Our method
maintains both hands, the object state, and the scene geometry.

\begin{figure*}[t]
  \centering
  \includegraphics[width=\textwidth]{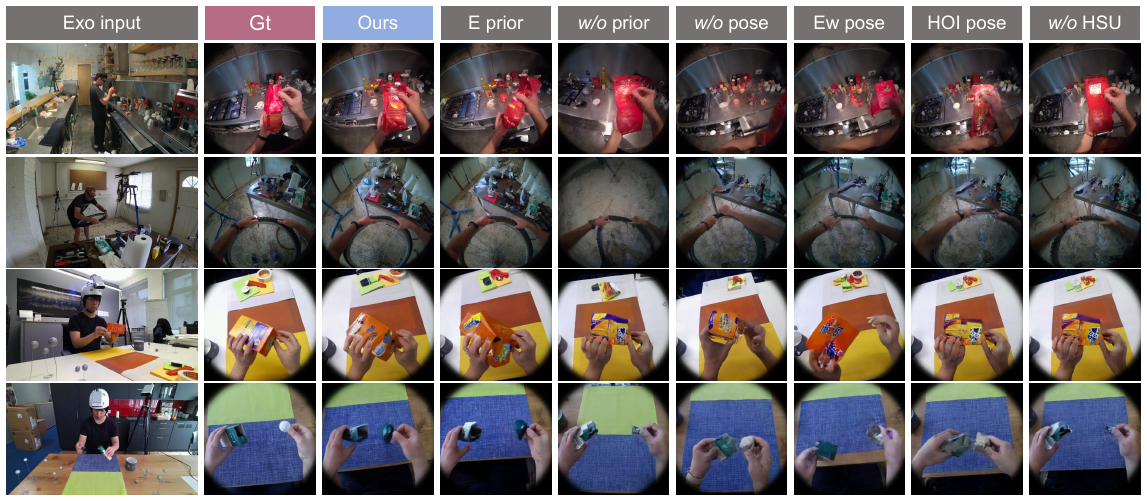}
  \caption{Four additional examples of the teacher ablations in Fig.~\ref{fig:ablation_comparison} on \cite{grauman2024egoexo4d} \textsc{Unseen}, covering the streaming dual-branch prior and region-aware loss weight scheduling (\textit{w/o} HSU). Each row shows one example. Gt accounts for egocentric ground truth.}
  \label{supp_fig:ablation_more}
\end{figure*}

Fig.~\ref{supp_fig:ablation_more} gives four additional examples of the teacher ablations in Fig.~\ref{fig:ablation_comparison}. Together with Fig.~\ref{fig:ablation_comparison} and the two scenes of Fig.~\ref{fig:ablation-student}, our qualitative ablations cover seven different examples.

\subsection{User study}

Following the evaluation protocol of prior work~\citep{kang2025egoxegocentricvideogeneration}, we conducted a blind user study to evaluate visual fidelity and perceptual coherence across challenging \textsc{Unseen} videos from \cite{grauman2024egoexo4d} and in-the-wild scenarios~\citep{h2o,sener2022assembly101}.
A total of 16 participants assessed shuffled, anonymized outputs generated by our method and competing baselines across 10 \textsc{Unseen} scene and 20 in-the-wild sequences.
For each comparative trial, participants were asked to select the best-performing result according to four criteria:
(1) \textbf{Overall Quality}: Which video delivers the highest perceptual realism and visual fidelity?
(2) \textbf{Viewpoint Consistency}: Which result best reflects the true target egocentric camera trajectory and perspective transfer?
(3) \textbf{Background Consistency}: Which method most faithfully reconstructs and preserves the observable background structure?
(4) \textbf{Hand-Object Interaction Accuracy}: Which model yields the most physically plausible and accurate articulated hand movements and manipulation dynamics?

\begin{wrapfigure}{!b}{0.5\textwidth}
\centering
\vspace{-15pt}
\includegraphics[width=\linewidth]{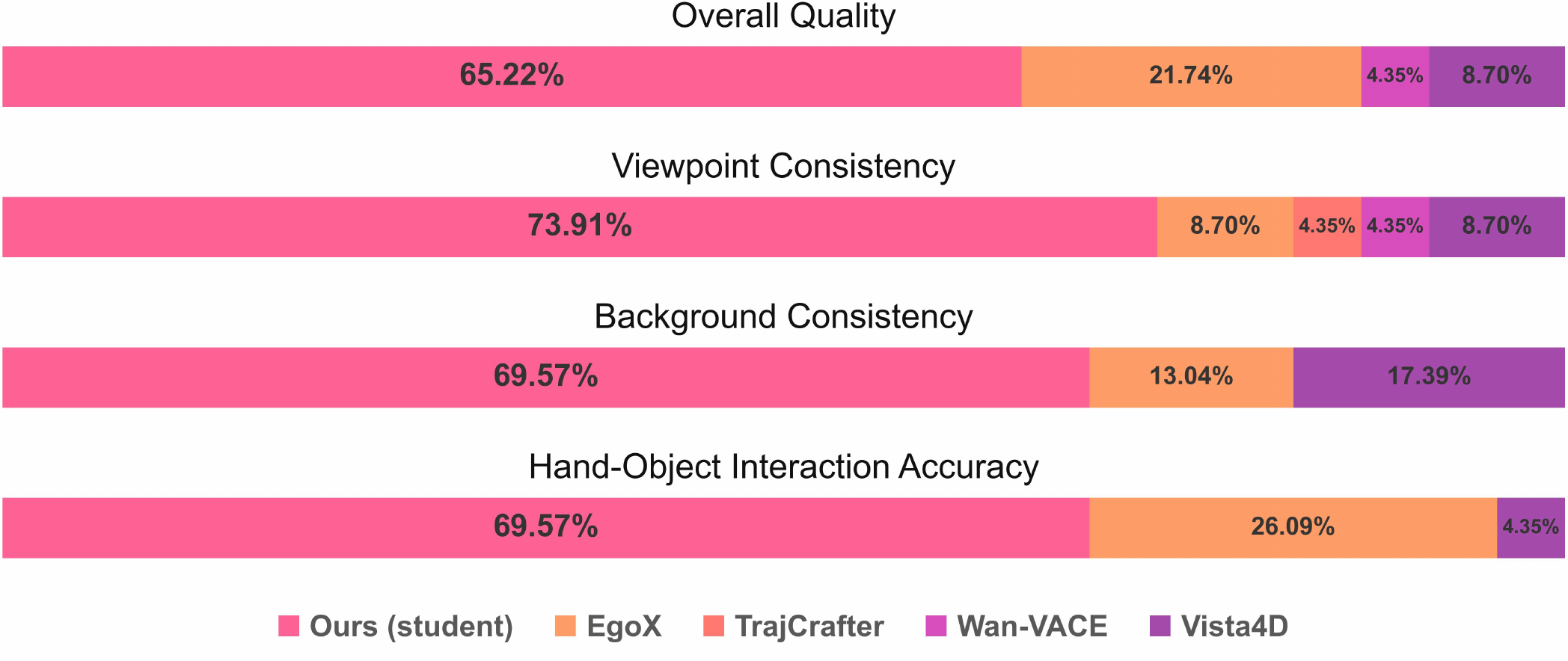}
\caption{Results for user study questions (1)-(4).}
\vspace{-15pt}
\label{fig:user_study}
\end{wrapfigure}

To further examine potential boundary artifacts and temporal drift inherent to chunk-wise causal generation, we designed a dedicated evaluation for video continuity:
(5) \textbf{Temporal Continuity}: Participants compared our streaming outputs against non-streaming baselines to evaluate whether noticeable stuttering, flickering, or abrupt transitions occur across chunk boundaries. Participants were asked to rate our results against offline methods~\citep{kang2025egoxegocentricvideogeneration} on temporal smoothness and the level of content/viewpoint drift:
5 points (highly consistent throughout with no perceptible drift),
4 points (minor but visually acceptable background drift),
and 3 points (moderate drift and flickers), 2 (noticeable drift and flickers that hinders perception) and 1 (severe drift and flickers).

As shown in Fig.~\ref{fig:user_study}, our approach received the majority of votes across all comparative dimensions, validating both its high fidelity in Exo2Ego transfer, interaction accuracy and cross-domain generalization. Furthermore, our streaming student scored an average of \textbf{4.6/5.0} on extended rollouts, corroborating that our chunk-wise causal architecture sustains long-horizon temporal stability.

\subsection{Variance over random seeds}
\label{sec:supp_seeds}

We re-run our models with five additional random seeds on identical inputs on all three test splits and report the mean\,$\pm$\,std in Table~\ref{tab:supp_seeds_main} ($\pm0.000$ denotes a std below $0.0005$). The ranking in Table~\ref{tab:short_video_comparison} holds, as the std is small compared with our margins over the baselines. Table~\ref{tab:supp_seeds_ablation} applies the same protocol to the ablations of Table~\ref{tab:ablation} (\textsc{Unseen}), where, as in Table~\ref{tab:ablation}, Full student is the 30-second rollout. The conclusions of Table~\ref{tab:ablation} hold as well.

\begin{table}[t]
  \centering
  \caption{Mean\,$\pm$\,std of our models over five additional random seeds on Ego-Exo4D~\citep{grauman2024egoexo4d} and H2O~\citep{h2o}.}
  \label{tab:supp_seeds_main}
  \renewcommand{\arraystretch}{1.4}
  \setlength{\tabcolsep}{3pt}
  \resizebox{\linewidth}{!}{%
  \begin{tabular}{llccccccccccccc}
    \toprule
    & Method & $\mathcal{S}$-PSNR $\uparrow$ & $\mathcal{S}$-MSSIM $\uparrow$ & $\mathcal{S}$-LPIPS $\downarrow$ & $\mathcal{U}$-CLIP-I $\uparrow$ & PCK@0.05 $\uparrow$ & MC $\uparrow$ & PSNR $\uparrow$ & SSIM $\uparrow$ & Loc Err $\downarrow$ & IoU $\uparrow$ & FVD $\downarrow$ & TF $\uparrow$ & MS $\uparrow$ \\
    \midrule
    \multirow{3}{*}{\rotatebox{90}{\shortstack{\textsc{New}\\\textsc{Action}}}} & Ours (teacher) & 18.30$\pm$0.13 & 0.606$\pm$0.002 & 0.365$\pm$0.004 & 0.904$\pm$0.003 & 51.95$\pm$2.31\% & 57.33$\pm$2.04\% & 17.62$\pm$0.09 & 0.600$\pm$0.001 & 0.068$\pm$0.006 & 0.483$\pm$0.009 & 107.40$\pm$3.48 & 0.983$\pm$0.000 & 0.994$\pm$0.000 \\
    & Ours (student) & 18.19$\pm$0.07 & 0.597$\pm$0.002 & 0.369$\pm$0.004 & 0.893$\pm$0.002 & 51.03$\pm$1.87\% & 55.63$\pm$1.95\% & 17.51$\pm$0.07 & 0.589$\pm$0.002 & 0.104$\pm$0.010 & 0.474$\pm$0.010 & 127.36$\pm$3.09 & 0.983$\pm$0.001 & 0.993$\pm$0.000 \\
    & Ours (long) & 18.02$\pm$0.09 & 0.588$\pm$0.002 & 0.372$\pm$0.006 & 0.890$\pm$0.005 & 45.68$\pm$2.35\% & 51.80$\pm$2.13\% & 17.43$\pm$0.07 & 0.588$\pm$0.001 & 0.127$\pm$0.013 & 0.450$\pm$0.015 & 139.18$\pm$6.49 & 0.980$\pm$0.001 & 0.991$\pm$0.000 \\
    \midrule
    \multirow{3}{*}{\rotatebox{90}{\textsc{Unseen}}} & Ours (teacher) & 18.24$\pm$0.14 & 0.579$\pm$0.003 & 0.412$\pm$0.008 & 0.897$\pm$0.003 & 43.45$\pm$2.69\% & 44.17$\pm$2.55\% & 17.07$\pm$0.10 & 0.547$\pm$0.002 & 0.093$\pm$0.010 & 0.453$\pm$0.014 & 133.85$\pm$4.12 & 0.981$\pm$0.000 & 0.995$\pm$0.000 \\
    & Ours (student) & 17.97$\pm$0.11 & 0.575$\pm$0.005 & 0.406$\pm$0.008 & 0.893$\pm$0.004 & 40.46$\pm$1.77\% & 41.35$\pm$2.12\% & 17.03$\pm$0.09 & 0.543$\pm$0.002 & 0.106$\pm$0.007 & 0.451$\pm$0.012 & 136.61$\pm$4.44 & 0.986$\pm$0.000 & 0.990$\pm$0.000 \\
    & Ours (long) & 17.75$\pm$0.10 & 0.572$\pm$0.003 & 0.401$\pm$0.012 & 0.884$\pm$0.005 & 36.26$\pm$2.80\% & 42.50$\pm$3.07\% & 16.97$\pm$0.09 & 0.534$\pm$0.003 & 0.121$\pm$0.014 & 0.436$\pm$0.011 & 151.02$\pm$4.98 & 0.989$\pm$0.001 & 0.993$\pm$0.001 \\
    \midrule
    \multirow{3}{*}{\rotatebox{90}{H2O}} & Ours (teacher) & 23.50$\pm$0.08 & 0.722$\pm$0.002 & 0.102$\pm$0.003 & 0.929$\pm$0.002 & 91.24$\pm$0.66\% & 95.31$\pm$0.36\% & 21.41$\pm$0.09 & 0.690$\pm$0.002 & 0.037$\pm$0.003 & 0.886$\pm$0.009 & 88.34$\pm$2.87 & 0.994$\pm$0.001 & 0.994$\pm$0.000 \\
    & Ours (student) & 23.43$\pm$0.08 & 0.714$\pm$0.003 & 0.109$\pm$0.004 & 0.926$\pm$0.002 & 85.26$\pm$0.87\% & 93.30$\pm$0.41\% & 21.04$\pm$0.11 & 0.685$\pm$0.003 & 0.052$\pm$0.003 & 0.854$\pm$0.011 & 90.01$\pm$2.75 & 0.993$\pm$0.000 & 0.994$\pm$0.000 \\
    & Ours (long) & 23.29$\pm$0.09 & 0.708$\pm$0.002 & 0.123$\pm$0.004 & 0.926$\pm$0.002 & 82.03$\pm$1.24\% & 92.65$\pm$1.32\% & 20.79$\pm$0.12 & 0.677$\pm$0.002 & 0.056$\pm$0.003 & 0.823$\pm$0.010 & 96.47$\pm$4.01 & 0.993$\pm$0.000 & 0.994$\pm$0.000 \\
    \bottomrule
  \end{tabular}}
\end{table}

\begin{table}[t]
  \centering
  \caption{Mean\,$\pm$\,std of the ablations of Table~\ref{tab:ablation} (\textsc{Unseen}) over five additional random seeds.}
  \label{tab:supp_seeds_ablation}
  \setlength{\tabcolsep}{3pt}
  \resizebox{\linewidth}{!}{%
  \begin{tabular}{lccccccccccc}
    \toprule
    Variant & $\mathcal{S}$-PSNR $\uparrow$ & $\mathcal{S}$-MSSIM $\uparrow$ & $\mathcal{S}$-LPIPS $\downarrow$ & $\mathcal{U}$-CLIP-I $\uparrow$ & PCK@0.05 $\uparrow$ & MC $\uparrow$ & Loc Err $\downarrow$ & IoU $\uparrow$ & FVD $\downarrow$ & TF $\uparrow$ & MS $\uparrow$ \\
    \midrule
    Full teacher & 18.24$\pm$0.14 & 0.579$\pm$0.003 & 0.412$\pm$0.008 & 0.897$\pm$0.003 & 43.45$\pm$2.69\% & 44.17$\pm$2.55\% & 0.093$\pm$0.010 & 0.453$\pm$0.014 & 133.85$\pm$4.12 & 0.981$\pm$0.000 & 0.995$\pm$0.000 \\
    w/o pose & 17.88$\pm$0.10 & 0.570$\pm$0.004 & 0.427$\pm$0.006 & 0.888$\pm$0.004 & 13.05$\pm$3.43\% & 21.05$\pm$3.17\% & 0.112$\pm$0.008 & 0.441$\pm$0.009 & 287.04$\pm$4.68 & 0.978$\pm$0.001 & 0.992$\pm$0.001 \\
    Ew pose & 18.16$\pm$0.11 & 0.570$\pm$0.003 & 0.415$\pm$0.006 & 0.891$\pm$0.004 & 19.34$\pm$3.22\% & 27.38$\pm$4.90\% & 0.103$\pm$0.010 & 0.443$\pm$0.010 & 281.67$\pm$3.41 & 0.980$\pm$0.000 & 0.988$\pm$0.000 \\
    HOI pose & 17.58$\pm$0.12 & 0.549$\pm$0.004 & 0.451$\pm$0.008 & 0.878$\pm$0.004 & 22.21$\pm$2.87\% & 23.06$\pm$3.01\% & 0.168$\pm$0.012 & 0.338$\pm$0.012 & 270.10$\pm$3.64 & 0.969$\pm$0.002 & 0.985$\pm$0.000 \\
    w/o prior & 13.49$\pm$0.14 & 0.394$\pm$0.003 & 0.606$\pm$0.008 & 0.716$\pm$0.005 & 15.16$\pm$4.42\% & 24.13$\pm$3.88\% & 0.242$\pm$0.011 & 0.126$\pm$0.014 & 662.81$\pm$7.10 & 0.958$\pm$0.001 & 0.962$\pm$0.000 \\
    E prior & 18.15$\pm$0.08 & 0.542$\pm$0.004 & 0.413$\pm$0.006 & 0.895$\pm$0.004 & 35.73$\pm$2.23\% & 41.02$\pm$2.34\% & 0.201$\pm$0.009 & 0.384$\pm$0.008 & 142.03$\pm$4.26 & 0.979$\pm$0.000 & 0.990$\pm$0.001 \\
    w/o HSU & 16.91$\pm$0.10 & 0.485$\pm$0.003 & 0.518$\pm$0.007 & 0.862$\pm$0.006 & 23.44$\pm$2.78\% & 32.90$\pm$2.31\% & 0.220$\pm$0.012 & 0.142$\pm$0.013 & 306.49$\pm$5.27 & 0.977$\pm$0.000 & 0.988$\pm$0.001 \\
    \midrule
    Full student & 17.75$\pm$0.10 & 0.572$\pm$0.003 & 0.401$\pm$0.012 & 0.884$\pm$0.005 & 36.26$\pm$2.80\% & 42.50$\pm$3.07\% & 0.121$\pm$0.014 & 0.436$\pm$0.011 & 151.02$\pm$4.98 & 0.989$\pm$0.001 & 0.993$\pm$0.001 \\
    w/o H corruption & 17.73$\pm$0.12 & 0.564$\pm$0.003 & 0.408$\pm$0.012 & 0.876$\pm$0.007 & 28.53$\pm$2.56\% & 36.29$\pm$2.24\% & 0.126$\pm$0.014 & 0.389$\pm$0.010 & 199.45$\pm$4.25 & 0.984$\pm$0.000 & 0.992$\pm$0.000 \\
    w/o first-chunk & 17.45$\pm$0.11 & 0.554$\pm$0.002 & 0.416$\pm$0.010 & 0.826$\pm$0.010 & 35.16$\pm$2.45\% & 39.23$\pm$3.31\% & 0.311$\pm$0.012 & 0.276$\pm$0.009 & 395.52$\pm$5.02 & 0.980$\pm$0.001 & 0.987$\pm$0.000 \\
    \bottomrule
  \end{tabular}}
\end{table}

\subsection{Drifting Score}
\label{sec:supp_drift}

We evaluate the drift of the student ablations with the Drifting Score of Helios~\citep{yuan2026helios}, using its official code, on the same 30-second rollouts on \textsc{Unseen} as Table~\ref{tab:ablation}. Each of its four metrics (Aesthetic, Smoothness, Semantic and Naturalness) is computed on the first and the last 15\% of the frames, and their absolute difference is mapped to an integer rating from 1 to 10 with the thresholds of Helios (higher means less drift). Since an egocentric video also changes by itself within 30 seconds, we score the ground-truth videos as a reference. We compare each end with the prompt of its own chunk, evaluate at our resolution of $480\times480$, and use GPT-6.1-sol~\citep{openai2026gpt61sol} as the Naturalness judge, so Naturalness is comparable across our rows but not with the values reported in Helios. As shown in Table~\ref{tab:supp_drift}, the full student stays close to the ground truth, while removing region-aware corruption or the first-chunk curriculum increases the drift.

\begin{table}[t]
  \centering
  \caption{Drifting Score~\citep{yuan2026helios} of the 30-second rollouts on \textsc{Unseen} (integer ratings from 1 to 10, higher means less drift).}
  \label{tab:supp_drift}
  \setlength{\tabcolsep}{6pt}
  \begin{tabular}{lcccc}
    \toprule
    Variant & Aesthetic $\uparrow$ & Smoothness $\uparrow$ & Semantic $\uparrow$ & Naturalness $\uparrow$ \\
    \midrule
    Ground truth & 7 & 10 & 7 & 10 \\
    \midrule
    Full student & 6 & 10 & 7 & 10 \\
    w/o H corruption & 6 & 9 & 7 & 9 \\
    w/o first-chunk & 4 & 8 & 7 & 7 \\
    \bottomrule
  \end{tabular}
\end{table}

\subsection{Text prompt at inference}
\label{sec:supp_text}

The online VLM observes only the current exocentric chunk ($6$ keyframes, Table~\ref{tab:supp_latency}), so it introduces no external information beyond the exocentric video. Moreover, PromptMix (Sec.~\ref{sec:supp_promptmix}) regularizes training to prevent the model from over-relying on text conditioning. To quantify the impact of inference-time captions versus the exocentric video and priors, we evaluate the student on \textsc{Unseen} across three caption detail levels generated by the same online VLM, namely (1) an empty string, (2) a description of the hand action and the manipulated object (ours), and (3) the manipulation description from (2) augmented with a detailed scene summary. In all variants, the VLM still identifies the dynamic objects for SAM~3.1~\citep{meta2026sam31} segmentation, isolating only the effect of the text prompt of the generator. Table~\ref{tab:supp_text} reports single runs, as in Table~\ref{tab:ablation}. The results confirm that visual appearance in $\mathcal{S}$ and $\mathcal{U}$ relies primarily on the exocentric video and the geometric priors rather than on text, whereas action semantics from the VLM provide supplementary guidance for hand-object interactions in $\mathcal{H}$. Adding detailed scene descriptions in (3) brings no further benefit over our concise prompt, showing that the model does not need scene captions from the VLM.

\begin{table}[t]
  \centering
  \caption{Student on \textsc{Unseen} with captions of different levels of detail at inference.}
  \label{tab:supp_text}
  \setlength{\tabcolsep}{3pt}
  \resizebox{\linewidth}{!}{%
  \begin{tabular}{lccccccccccc}
    \toprule
    Text prompt at inference & $\mathcal{S}$-PSNR $\uparrow$ & $\mathcal{S}$-MSSIM $\uparrow$ & $\mathcal{S}$-LPIPS $\downarrow$ & $\mathcal{U}$-CLIP-I $\uparrow$ & PCK@0.05 $\uparrow$ & MC $\uparrow$ & Loc Err $\downarrow$ & IoU $\uparrow$ & FVD $\downarrow$ & TF $\uparrow$ & MS $\uparrow$ \\
    \midrule
    (1) Empty & 17.86 & 0.575 & 0.415 & 0.866 & 26.60\% & 30.97\% & 0.131 & 0.413 & 173.53 & 0.988 & 0.993 \\
    (2) Concise (ours) & 17.88 & 0.569 & 0.394 & 0.890 & 37.22\% & 41.70\% & 0.119 & 0.427 & 144.33 & 0.988 & 0.994 \\
    (3) Detailed & 17.91 & 0.572 & 0.390 & 0.894 & 37.51\% & 39.01\% & 0.127 & 0.423 & 147.65 & 0.988 & 0.993 \\
    \bottomrule
  \end{tabular}}
\end{table}

\subsection{Backbone}
\label{sec:supp_backbone}

To ablate the backbone, we port EgoX~\citep{kang2025egoxegocentricvideogeneration} from Wan2.1-I2V-14B to our Wan2.2-TI2V-5B backbone~\citep{team2025wan}, keeping its original ego prior, conditioning, and GGA unchanged. We fine-tune it on our training set under the same resolution, clip length, and frame rate as our model, and evaluate it on the same protocol as Table~\ref{tab:short_video_comparison}. Table~\ref{tab:supp_backbone} also lists the original EgoX on Wan2.1-I2V-14B and Ours (teacher) from Table~\ref{tab:short_video_comparison}. On the same backbone as ours, EgoX performs on par with its original version and remains far below ours, which confirms that the backbone does not account for the performance gap in Table~\ref{tab:short_video_comparison}. The effect of the prior and segmentation method is ablated by the E prior row of Table~\ref{tab:ablation}, which replaces our dynamic ego prior (using SAM~3.1, \citealp{meta2026sam31}) with the offline ego prior of EgoX (using SAM-ViT-B, \citealp{kirillov2023segment}) while keeping the same human pose prior.

\begin{table}[t]
  \centering
  \caption{EgoX~\citep{kang2025egoxegocentricvideogeneration} on its original backbone and on ours, evaluated on Ego-Exo4D~\citep{grauman2024egoexo4d} and H2O~\citep{h2o}.}
  \label{tab:supp_backbone}
  \renewcommand{\arraystretch}{1.4}
  \setlength{\tabcolsep}{3pt}
  \resizebox{\linewidth}{!}{%
  \begin{tabular}{lllcccccccccccccc}
    \toprule
    & Method & Backbone & $\mathcal{S}$-PSNR $\uparrow$ & $\mathcal{S}$-MSSIM $\uparrow$ & $\mathcal{S}$-LPIPS $\downarrow$ & $\mathcal{U}$-CLIP-I $\uparrow$ & PCK@0.05 $\uparrow$ & MC $\uparrow$ & PSNR $\uparrow$ & SSIM $\uparrow$ & Loc Err $\downarrow$ & IoU $\uparrow$ & FVD $\downarrow$ & TF $\uparrow$ & MS $\uparrow$ & FPS $\uparrow$ \\
    \midrule
    \multirow{3}{*}{\rotatebox{90}{\shortstack{\textsc{New}\\\textsc{Action}}}} & EgoX & Wan2.1-I2V-14B (orig.) & 15.75 & 0.489 & 0.580 & 0.888 & 25.91\% & 34.84\% & 16.06 & 0.561 & 0.130 & 0.357 & 184.24 & 0.977 & 0.990 & 0.08 \\
    & EgoX & Wan2.2-TI2V-5B & 16.01 & 0.505 & 0.575 & 0.873 & 27.06\% & 32.42\% & 16.31 & 0.573 & 0.135 & 0.339 & 195.26 & 0.974 & 0.990 & 0.46 \\
    & Ours (teacher) & Wan2.2-TI2V-5B & 18.35 & 0.603 & 0.372 & 0.907 & 49.86\% & 56.26\% & 17.51 & 0.603 & 0.076 & 0.478 & 111.97 & 0.983 & 0.994 & 0.97 \\
    \midrule
    \multirow{3}{*}{\rotatebox{90}{\textsc{Unseen}}} & EgoX & Wan2.1-I2V-14B (orig.) & 13.36 & 0.406 & 0.577 & 0.879 & 11.07\% & 15.41\% & 14.41 & 0.456 & 0.311 & 0.092 & 438.79 & 0.981 & 0.992 & 0.08 \\
    & EgoX & Wan2.2-TI2V-5B & 13.59 & 0.412 & 0.554 & 0.888 & 12.31\% & 15.23\% & 14.86 & 0.448 & 0.297 & 0.127 & 468.69 & 0.988 & 0.990 & 0.46 \\
    & Ours (teacher) & Wan2.2-TI2V-5B & 18.18 & 0.576 & 0.405 & 0.901 & 40.24\% & 43.76\% & 17.19 & 0.544 & 0.106 & 0.456 & 138.91 & 0.980 & 0.995 & 0.97 \\
    \midrule
    \multirow{3}{*}{\rotatebox{90}{H2O}} & EgoX & Wan2.1-I2V-14B (orig.) & 14.23 & 0.465 & 0.377 & 0.710 & 35.10\% & 40.28\% & 15.00 & 0.468 & 0.205 & 0.513 & 479.52 & 0.911 & 0.929 & 0.08 \\
    & EgoX & Wan2.2-TI2V-5B & 14.45 & 0.476 & 0.359 & 0.694 & 38.58\% & 39.37\% & 15.52 & 0.490 & 0.186 & 0.505 & 501.25 & 0.928 & 0.939 & 0.46 \\
    & Ours (teacher) & Wan2.2-TI2V-5B & 23.59 & 0.724 & 0.106 & 0.931 & 89.22\% & 95.78\% & 21.35 & 0.688 & 0.041 & 0.876 & 89.94 & 0.993 & 0.994 & 0.97 \\
    \bottomrule
  \end{tabular}}
\end{table}

\section{End-to-End Latency}
\label{sec:supp_latency}

\noindent\textbf{Setup.} We analyze the runtime of the deployed streaming
student. Model loading---transferring weights to the GPU and initializing the
prior estimators---is a one-time cost and is excluded from all figures below.
The student consumes one chunk at a time; a chunk spans $L=33$ video frames at
$24$\,fps, \textit{i.e.}\ $1.375$\,s of video. Each stage is measured on the
device stated in Table~\ref{tab:supp_latency}. Our end-to-end latency is measured on H100 GPU.

\begin{figure}[tbp]
\centering
\includegraphics[width=\textwidth]{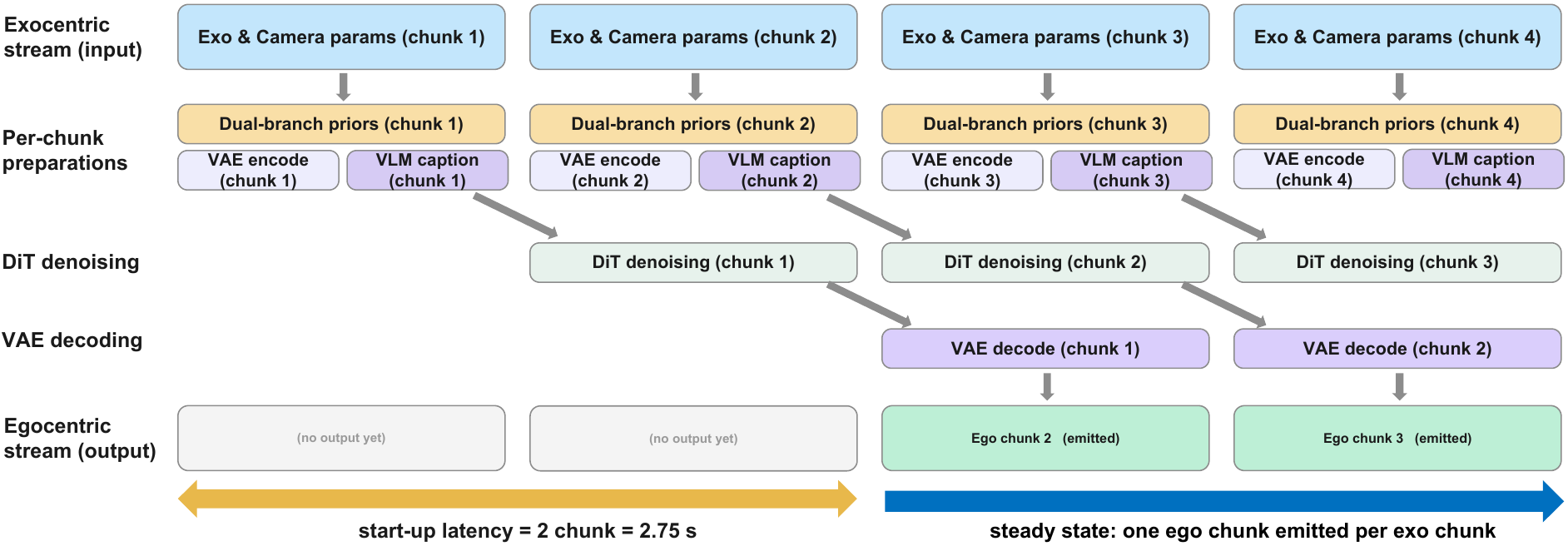}
\caption{Streaming schedule of the student. Each incoming exocentric chunk
triggers a fixed sequence of stages; the corresponding egocentric chunk is
emitted when the \emph{next} exocentric chunk arrives, giving a constant
start-up latency of one chunk ($1{,}375$\,ms).}
\label{fig:supp_latency}
\end{figure}

\noindent\textbf{Per-chunk compute.} Within a chunk the stages are ordered by
data dependency, and the order matters for a streaming implementation. Building
the two priors comes first: the dynamic ego prior requires depth, segmentation,
and cache updates, while the human pose prior requires body and hand
estimation. The two are independent branches and run concurrently, so the stage
costs the time of the slower branch. \emph{Both are prerequisites for their own
VAE encoding}, and the encoded priors in turn must be available before the
diffusion transformer can denoise, since they enter the model as pixel-aligned
conditioning channels. Encoding of the exocentric chunk and of the prior
streams is independent and is therefore issued as one batched call. VLM
inference depends only on the raw exocentric frames, so it is independent of
the priors and the VAE, and runs alongside them. Decoding depends on denoising
and follows it. Table~\ref{tab:supp_latency} reports the measured cost of each
stage.

\begin{table}[t]
  \centering
  \caption{Per-chunk stage costs of the streaming student ($33$ frames,
  $1{,}375$\,ms of video). Stages in the same \emph{parallel group} are
  independent and can be executed concurrently; the dependency column states
  what each stage must wait for.}
  \label{tab:supp_latency}
  \renewcommand{\arraystretch}{1.05}
  \setlength{\tabcolsep}{4pt}
  \begin{tabular}{llcrr}
    \toprule
    Stage & Device & Depends on & Par. & Latency \\
    \midrule
    (1) Streaming dual-branch priors & H100 & --- & A & $0.58$\,s \\
    (2) VAE encode, exo chunk + priors & H100 & (1) & B & $0.30$\,s \\
    (3) VLM action prompt ($6$ keyframes) & H100 & --- & B & $0.23$\,s \\
    (4) Denoising, $3$ steps (student) & H100 & (2), (3) & C & $1.36$\,s \\
    (5) VAE decode, ego chunk & H100 & (4) & D & $0.91$\,s \\
    \bottomrule
  \end{tabular}
\end{table}

\noindent\textbf{Execution pipeline.}
Rather than executing stages serially, our system pipelines conditioning, generation, and decoding asynchronously across chunks (Fig.~\ref{fig:supp_latency}).
When exocentric chunk $k$ arrives, its dual-branch priors and VAE latents are prepared concurrently; because the VLM branch runs faster than VAE encoding ($0.23$\,s vs. $0.58$\,s), its latency is entirely hidden.
Once chunk $k{+}1$ arrives, the student denoises chunk $k$ conditioned on these precomputed priors and historical context to produce clean egocentric latents ($1.36$\,s).
Finally, as chunk $k{+}2$ arrives, the VAE decodes the clean latents into RGB frames ($0.91$\,s).
Because individual stages execute within the chunk duration ($1.375$\,s) and run in parallel, steady-state throughput sustains interactive frame rates, introducing an end-to-end streaming latency of exactly two chunks.

This is what makes the few-step distillation of Sec.~\ref{sec:method_student}
necessary rather than merely convenient. The bidirectional teacher requires
$50$ steps and runs at $0.97$\,fps, so a single chunk would need roughly
$34$\,s and no amount of pipelining could hide it. Distilling the student to
$3$ steps raises the generation throughput to $24.3$\,fps on one H100, which is
the figure reported in Table~\ref{tab:short_video_comparison}.

\section{Implementation Details}
\label{sec:supp_implementation}

\subsection{Streaming dual-branch priors}
\label{sec:supp_priors}

Both priors of Sec.~\ref{sec:dual_condition} are chunk-causal
estimators rather than offline preprocessing: each consumes the exocentric
chunk that has just arrived together with a bounded internal state and emits
exactly the prior frames the current chunk requires. The dynamic ego prior
exploits the fact that the exocentric camera is stationary, so every pixel of
the rectified exocentric image corresponds to a \emph{fixed ray in world
space}, and only which ray observes a surface changes over time. We therefore
keep two memories with different lifetimes: a ray-indexed static cache
$\mathcal{A}_t = \{d^A_t(p), c^A_t(p), s_t(p)\}$ holding the first depth and
colour observed along each ray, and a short-lived dynamic state holding the
previous frame's masks, depth, and flow reference. The cache holds
$|\Omega_e|$ entries ($624\times360$ in deployment) irrespective of stream
length, and is written only at keyframes, and only where the observation is
valid, free of dynamics, and new: background revealed after an occlusion is
filled in when it becomes observable, and an accepted entry is never revisited,
so the static branch accumulates no recursive drift---though the same rule
freezes the first estimate and offers no way to invalidate it. Segmentation and
monocular depth run on keyframes only (every $12$ frames), and intervening
frames propagate the previous state along backward optical flow, so depth is
refreshed from a genuine estimate at most $11$ frames old. Depth and masks are
propagated, but the \emph{colour} of dynamic surfaces always comes from the
current frame, so hands and manipulated objects keep current appearance, and
dynamic points are never written into $\mathcal{A}$. Static and dynamic
surfaces are then projected into the target camera and resolved in a single
depth buffer, with source points reconstructed in the kernel from ray and depth
so that no point cloud reaches the host; projection follows the calibrated Aria
Fisheye624 model, each point estimates a local footprint radius to suppress
striping on grazing surfaces, and the two surface types compete on equal terms
so that occlusion is respected rather than the dynamic layer being composited
on top of the background.

The human pose prior supplies the structural guidance that pixel-aligned
geometry cannot, since it does not by itself disambiguate which arm is which or
how fingers are articulated. As estimating hands directly in the exocentric
frame is unreliable---the regions are small and hands cross---we bond the hand
to the body. Per sampled frame, SMPLer-X~\citep{cai2023smpler} estimates the
body and ViTPose~\citep{xu2022vitpose} refines the upper-limb joints, with the
principal subject selected using the projection of the \emph{given} egocentric
camera centre and therefore no egocentric pixels. The wrist $B^{s}_{t,\mathrm{wrist}}$ of Eq.~\ref{eq:body_guided_hand} is obtained in the calibrated exocentric camera frame in three steps. (1) \emph{Metric anchoring}: we discard the unreliable camera-relative translation of SMPLer-X, retaining only its head-relative 3D offsets $\widehat{B}_t-\widehat{B}_{t,\mathrm{head}}$, and anchor the head to the calibrated egocentric camera centre $c_t$ of $T_{g,t}$, already given as input in Eq.~\ref{eq:dynamic_render}. (2) \emph{Scale and offset fitting}: we parameterize the metric 3D body joints as $B_t = \gamma\,(\widehat{B}_t-\widehat{B}_{t,\mathrm{head}}) + c_t + \Delta$ and solve for the global metric scale $\gamma$ and the small head-to-camera offset $\Delta$ by minimizing the 2D reprojection error against the ViTPose detections under the calibrated exocentric intrinsics. (3) \emph{Arm and wrist refinement}: the arm joints are refined identically under bone-length constraints, and a high-confidence WiLoR~\citep{potamias2025wilor} detection further pulls the wrist toward the detected 2D wrist location while preserving its fitted metric depth. Because $B_t$ resides directly in the calibrated metric camera space, $B^{s}_{t,\mathrm{wrist}}$ and the resulting hand joints $J^{s}_{t,j}$ are projected into both views without any egocentric pixels or inputs beyond Eq.~\ref{eq:dynamic_render}. Detections are associated to
the left and right wrists by a size-normalized linear assignment, which
enforces anatomical sidedness even when arms cross or the subject turns, and
the resulting wrist joints define the crop boxes for
WiLoR~\citep{potamias2025wilor}, recovering wrist-relative finger articulation;
when no hand is detected, a proposal along the forearm carries a deliberately
low confidence rather than being silently dropped. Predicted fingers are then
rigidly anchored so the hand root coincides with the refined wrist, as in
Eq.~\ref{eq:body_guided_hand}, fixing global position from the arm chain while
preserving local articulation---the kinematic analogue of proprioception, and
what prevents the spatial misalignment that isolated hand estimation produces on
the egocentric canvas. Since only a few frames per chunk are estimated, the rest
are interpolated, and a two-bone IK step re-solves each elbow afterwards to
restore the limb lengths that interpolation perturbs. The skeleton is projected
into both views with the calibrated camera parameters to yield
$(Q^e_t, Q^g_t)$, with joint confidence modulating stroke intensity so that
speculative finger poses are not presented at full contrast.

Because the cache is $\mathcal{O}(|\Omega_e|)$ and the expensive networks run
only on a few keyframes per chunk, the cost of a chunk does not grow with the
number of chunks already processed, and the two independent branches can execute
concurrently, so the stage costs the time of the slower one.
The combined prior stage enters the end-to-end budget as the $0.58$\,s of
Table~\ref{tab:supp_latency}.

\subsection{Bidirectional Teacher: PromptMix}
\label{sec:supp_promptmix}

The bidirectional teacher receives dense visual conditions from the exocentric
video and from the two streaming priors. If the text prompt is allowed to
describe scene appearance, it becomes a shortcut: the model can satisfy the
training objective by copying environmental wording into the egocentric canvas
instead of reading the pixel-aligned prior, and small changes in that wording
then alter the output even when the visual evidence is unchanged. PromptMix
controls how much information is available through cross-attention by sampling,
per training clip, one of four prompt variants:

\begin{itemize}
  \item a \emph{concise} description of the hand action and the manipulated
  object, with appearance adjectives (colours, materials, brands) stripped;
  \item a \emph{detailed exocentric} description that adds a scene summary and
  richer exocentric context;
  \item a \emph{full} description combining both;
  \item an \emph{empty} string.
\end{itemize}

All variants are derived from exocentric annotations; none contains a
description of the ground-truth egocentric frame. The variant is sampled per
clip with probabilities $0.3$, $0.5$, $0.1$, and $0.1$ respectively, and the
concise variant is used at validation and inference. The empty branch is a text
conditioning dropout and is not used in the classifier-free-guidance sense: we
sample with a guidance scale of $1.0$ and never evaluate an unconditional
branch.

The design intent is that the visual streams---above all the pixel-aligned
dynamic ego prior---become the primary route by which scene layout and
appearance reach the model, while text supplies only coarse action semantics.
PromptMix is a regularizer rather than a mechanism with an expected accuracy
gain.

\subsection{Streaming Adaptation}
\label{sec:supp_ar_transition}

We adapt the bidirectional teacher into a causal, chunk-wise autoregressive
student in three stages, following Helios~\citep{yuan2026helios}.

\noindent\textbf{Overview.} \emph{Stage 1} converts the bidirectional backbone
into an autoregressive generator by introducing causal attention across chunks.
\emph{Stage 2} reduces the number of noisy tokens per step through a spatial
pyramid, which is what makes the per-chunk cost tractable.
\emph{Stage 3} distills the multi-step generator into a few-step student
through distribution matching and adversarial training. The region-aware
corruption mechanism and the first-chunk curriculum of
Sec.~\ref{sec:method_student} are both introduced in Stage 1 and are retained
through Stages 2 and 3. Table~\ref{tab:supp_ar_hyper} lists our training
configuration of Stages~1 and~2; Table~\ref{tab:supp_s3_hyper} details Stage 3.

\noindent\textbf{Stage 1: causal adaptation.} The student warm-starts from the
bidirectional teacher. Attention
remains bidirectional \emph{within} a chunk while being strictly causal across
chunks. A $121$-frame clip encodes to $31$ latents, which we partition into
chunks spanning latents $(0,9)$, $(9,18)$, $(18,27)$, and $(27,31)$; the last
chunk is shorter and its loss is rescaled by its length fraction so that it
does not dominate the average.

The memory buffer stores egocentric latents only. The exocentric video and both
priors are supplied fresh for the current chunk and are \emph{discarded} rather
than accumulated, so the conditioning cost does not grow with rollout length;
distant history is compressed by multi-term memory patchification~\citep{yuan2026helios} into short-,
mid-, and long-term scales with histories of $(16,2,1)$ latent frames.
Annotations marking prior validity and dynamic regions accompany the cached
latents.

Two departure points from a global prompt are worth noting. First, we use a
\emph{per-chunk} prompt describing the action in the chunk currently being
denoised, rather than one prompt for the whole clip; text cross-attention is
restricted to the current chunk's tokens, so the current description is never
re-applied to cached history. This matches the teacher's conditioning
semantics---a description of the target being denoised---and allows the action
language to follow a changing manipulation without leaking future-chunk text
into earlier generations. Second, we regularize the conditioning path
throughout training: with probability $0.1$ the prior is dropped and with
probability $0.1$ it is corrupted with Gaussian noise ($\sigma = 0.25$),
applied only to the egocentric half of the conditioning channels, which is
always zeroed for the exocentric half.

Stage 1 is trained in three sub-phases, each re-loading the previous stage's
terminal checkpoint with a fresh optimizer. The sequence implements the
first-chunk curriculum and then the region-aware corruption of
Sec.~\ref{sec:method_student}: sub-phase A trains on the history-free first
chunk only (sampling probability $1.0$); sub-phase B reduces that probability
to $0.6$ and admits continuation chunks; sub-phase C lowers it to $0.25$ and
switches on history corruption. Corruption applies to history frames only, and
never to the first chunk, which has no history. It combines a frame-level
operator (clean $0.3$ / blur $0.6$ / noise $0.1$, the latter mixing with
$\sigma \sim \mathcal{U}(0, 1/3)$) with a targeted operator restricted to the
hand and interaction masks on the latent grid, which fires per frame with
probability $\sim \mathcal{U}(0.5, 0.7)$ and, on firing, applies either
stronger noise ($\sigma \sim \mathcal{U}(0.2, 0.5)$) or stronger blur. The
targeted operator forces the student to resolve hand dynamics from the current
priors rather than from a degraded history.

\begin{table}[tb]
  \centering
  \caption{Configuration of the causal adaptation (Stage~1) and of the pyramid
  token compression (Stage~2). Each sub-phase re-initializes the optimizer.
  $\dagger$\,The Stage-1 sub-phases share a single LoRA configuration.}
  \label{tab:supp_ar_hyper}
  \renewcommand{\arraystretch}{1.02}
  \setlength{\tabcolsep}{4pt}
  \resizebox{\linewidth}{!}{%
  \begin{tabular}{lccccc}
    \toprule
    Configuration & Stage-1 A & Stage-1 B & Stage-1 C & Stage-2 init & Stage-2 post \\
    \midrule
    Initialization & teacher & 1A & 1B & 1C & 2 init \\
    Training steps & $2{,}500$ & $10{,}000$ & $18{,}000$ & $16{,}000$ & $20{,}000$ \\
    Learning rate & $1\!\times\!10^{-5}$ & $1\!\times\!10^{-5}$ & $6\!\times\!10^{-6}$ & $2\!\times\!10^{-5}$ & $6\!\times\!10^{-6}$ \\
    LR schedule & constant w/ $100$-step warmup & $\leftarrow$ & $\leftarrow$ & $\leftarrow$ & $\leftarrow$ \\
    First-chunk probability & $1.0$ & $0.6$ & $0.25$ & $0.25$ & $0.25$ \\
    History corruption & off & off & on & on & on \\
    Pyramid sample ratios & --- & --- & --- & $(1,2,1)$ & $(1,1,1)$ \\
    Global batch size & $8$ & $8$ & $8$ & $8$ & $8$ \\
    Optimizer & \multicolumn{5}{c}{AdamW, $\beta_1{=}0.9$, $\beta_2{=}0.95$, $\epsilon{=}10^{-8}$, weight decay $10^{-4}$} \\
    Gradient clipping & $1.0$ & $1.0$ & $1.0$ & $1.0$ & $1.0$ \\
    LoRA rank / $\alpha$ (q, k, v, o)$^\dagger$ & $256 / 256$ & $\leftarrow$ & $\leftarrow$ & $\leftarrow$ & $\leftarrow$ \\
    LoRA rank / $\alpha$ (FFN) & $128 / 128$ & $\leftarrow$ & $\leftarrow$ & $\leftarrow$ & $\leftarrow$ \\
    Trainable modules & \multicolumn{5}{c}{LoRA adapters, patch embedding, memory patchifiers} \\
    Numerical precision & bf16 & bf16 & bf16 & bf16 & bf16 \\
    GPU usage & 8$\times$H800 & 8$\times$H800 & 8$\times$H800 & 8$\times$H800 & 8$\times$H800 \\
    \bottomrule
  \end{tabular}}
\end{table}

\noindent\textbf{Stage 2: pyramid token compression.} Denoising is divided into
three spatial levels while each chunk retains its full temporal span. At
$832\times480$ exocentric and $480\times480$ egocentric resolution the two
views occupy different widths on the latent grid, so they are resized
separately; interpolation is never allowed to blend pixels across the seam
between them. At every level the exocentric half is clamped to its clean
conditioning latent and the flow-matching target and loss apply to the
egocentric half only. Training first emphasizes the intermediate level with
sampling ratios $(1,2,1)$, then balances the levels with $(1,1,1)$ at a lower
learning rate. Because the finest latent grid is not divisible by successive
factors of two, we sample independent unit-variance Gaussian noise directly at
each level rather than downsampling a single fine noise tensor, which would
otherwise produce spatially varying variance; the finest transition
additionally uses negatively correlated block noise to compensate for the
correlation introduced by nearest-neighbor upsampling.

\noindent\textbf{Stage 3: few-step distillation.} Stage 3 removes the remaining
inference cost. It proceeds in two sub-phases. The first is an ODE
initialization: the Stage-2 model is treated as a frozen teacher, and a compact
set of ODE solution pairs is collected from it, which the student fits by
regression at all pyramid levels simultaneously. The second is full
adversarial distillation, which follows a generalized Distribution Matching
Distillation with an adversarial term (Eq.~\ref{eq:ar_distillation}). We
instantiate the real- and fake-score estimators from frozen teacher copies; the
fake-score branch additionally carries trainable critic parameters and
discriminator heads, and since the two estimators are identical at
initialization, the critic must depart from the frozen teacher before it can
supply a non-degenerate signal. Only the finest pyramid level is scored.
Classifier-free guidance is disabled throughout---both estimators are
evaluated conditionally---which removes the extra forward pass and closes the
train/inference gap. The generator is updated once every five critic updates,
and the discriminator heads are attached at four intermediate depths plus the
output. Configurations are listed in Table~\ref{tab:supp_s3_hyper}.

\begin{table}[tb]
  \centering
  \caption{Configuration of the few-step distillation (Stage~3).}
  \label{tab:supp_s3_hyper}
  \setlength{\tabcolsep}{5pt}
  \begin{tabular}{lcc}
    \toprule
    Configuration & Stage-3 ODE init & Stage-3 post \\
    \midrule
    Initialization & Stage-2 post & Stage-3 ODE init \\
    Training steps & $3{,}759$ & $2{,}250$ \\
    Generator learning rate & $2\!\times\!10^{-6}$ & $2\!\times\!10^{-6}$ \\
    Critic learning rate & --- & $4\!\times\!10^{-7}$ \\
    LR schedule & constant & constant \\
    Optimizer & AdamW, $\beta_1{=}0$, $\beta_2{=}0.999$, wd $10^{-3}$ & $\leftarrow$ \\
    Gradient clipping (generator / critic) & $1.0$ / --- & $1.0$ / $10.0$ \\
    ODE regression weight & $80.0$ & --- \\
    Adversarial weight $\lambda_G$ (generator) & --- & $5\!\times\!10^{-2}$ \\
    Adversarial weight $\lambda_D$ (critic) & --- & $1\!\times\!10^{-2}$ \\
    R1 weight / $\sigma$ & --- & $100.0$ / $0.1$ \\
    Critic update ratio (TTUR) & --- & $5$ \\
    Discriminator start step & --- & $1{,}000$ \\
    Denoising steps per level & $2$ & $2$ \\
    Real-/fake-score guidance scale & $1.0$ / $1.0$ & $1.0$ / $1.0$ \\
    Numerical precision & bf16 & bf16 \\
    GPU usage & 4$\times$A800 & 4$\times$A800 \\
    \bottomrule
  \end{tabular}
\end{table}

\subsection{Bidirectional Teacher Training}
\label{sec:supp_teacher_training}

The bidirectional teacher is initialized from Wan2.2-TI2V-5B~\citep{team2025wan}
and trained for $34{,}000$ steps on $8$ H800 GPUs with a global batch size of
$8$, at a constant learning rate of $2\!\times\!10^{-5}$, in bf16, with AdamW
and ZeRO-2. LoRA ($r{=}\alpha{=}256$) is applied to the attention projections.
Clips span $121$ frames at $24$\,fps, with exocentric inputs at $832\times480$
and egocentric targets at $480\times480$. PromptMix probabilities are
$0.3/0.5/0.1/0.1$ as in Sec.~\ref{sec:supp_promptmix}, and the pose condition
is dropped with probability $0.05$.

The two conditioning branches enter the backbone as additional input channels
of the patch embedding, concatenated with the noisy latent along the channel
dimension. Because the backbone is warm-started from a pretrained model whose
noisy-latent channels are already well scaled, the new branches are initialized
as scaled copies of that block rather than from scratch; otherwise a randomly
initialized branch would dominate the pretrained channels at step zero and
destroy the warm start. We set the initialization scale of the dynamic ego
prior branch to $0.5$ and that of the human pose prior branch to $0.19$ of the
noisy-latent block. The scale of the pose branch is deliberately smaller than
that of the appearance branch: the pose prior is a sparse, high-contrast
skeleton map, and at parity it would inject a far larger activation than the
dense appearance prior.

\subsection{Evaluation Metrics}
\label{sec:supp_metrics}

\noindent\textbf{Why region-aware metrics.} A conventional pixel-level
comparison over the whole egocentric canvas is the wrong instrument for this
task, because the three regions have different correctness criteria. In
$\mathcal{S}$ the view is shared with the exocentric camera, and the generated
content must match the ground truth essentially exactly; a pixel metric is
meaningful there. In $\mathcal{U}$ no observation exists by definition, so
requiring pixel agreement with the ground truth penalizes legitimate and
desirable diversity: a different but coherent completion of an unseen region is
not an error. In $\mathcal{H}$ the situation is subtler. The hand and object
are not always observable, and their egocentric appearance is not pixel-determined---a
slightly different wrist roll, a shifted contact point, or a slightly different
grasp all produce plausible images with low pixel similarity to the recorded
frame. What matters in $\mathcal{H}$ is \emph{semantic} agreement: the action
performed, the positions of the hands, and the spatial relation between hand
and object.

\noindent\textbf{Region definition.} We derive $\mathcal{H}$--$\mathcal{S}$--$\mathcal{U}$ region masks by combining geometric projection with segmentation on ground-truth egocentric videos. Following the streaming geometric construction in Sec.~\ref{sec:dual_condition}, we build a 3D world cache from the exocentric stream and project the cached points into each target egocentric view using the calibrated camera parameters. Pixels covered by valid projections form the initial shared-view region $\mathcal{S}'$, while the remaining pixels form the initial unseen-view region $\mathcal{U}'$. We then apply SAM~3.1~\citep{meta2026sam31} to the ground-truth egocentric video, using words from text prompts to segment and track dynamic objects and human limbs; the union of these masks defines the interaction region $\mathcal{H}$. To resolve overlaps, we assign all pixels in $\mathcal{H}$ to the interaction region and set $\mathcal{S}=\mathcal{S}'\setminus\mathcal{H}$ and $\mathcal{U}=\mathcal{U}'\setminus\mathcal{H}$. The resulting three regions are mutually exclusive and together cover the entire egocentric image. These masks support the region-aware loss curriculum during training and region-specific metric computation during evaluation. They are not required at inference time, and generation does not require access to ground-truth egocentric videos.

\noindent\textbf{Image metrics.} \emph{$\mathcal{S}$-PSNR} is the PSNR
restricted to $\mathcal{S}$, with the mean squared error normalized by the
number of pixels inside the mask rather than by the full canvas, so that
small-region scores are not inflated. \emph{$\mathcal{S}$-MSSIM} applies SSIM
with an $11\times11$ Gaussian window ($\sigma{=}1.5$) whose local moments are
renormalized by the in-mask Gaussian weight, so that pixels outside
$\mathcal{S}$ never enter the local statistics; it therefore returns exactly
$1.0$ for a prediction that differs from the ground truth only in
$\mathcal{U}$ or $\mathcal{H}$. \emph{$\mathcal{S}$-LPIPS} uses the AlexNet
LPIPS distance map, pooled with the same $\mathcal{S}$ mask; note that LPIPS is
evaluated on the full frame pair, so its receptive field includes context
outside $\mathcal{S}$ and the score is $\mathcal{S}$-weighted rather than
$\mathcal{S}$-isolated.

\emph{$\mathcal{U}$-CLIP-I} measures the naturalness and visual plausibility of
the unseen region by CLIP image-embedding similarity between corresponding
predicted and ground-truth frames. We deliberately use a similarity
(plausibility) measure here rather than a fidelity measure, for the reason
given above: $\mathcal{U}$ is unobservable, so the standard is coherence, not
agreement with one realization.

\emph{PCK@0.05} measures hand pose accuracy in $\mathcal{H}$. Hand keypoints
are detected independently in the prediction and in the ground truth using \cite{yang2012articulated}, matched hand-to-hand by a Hungarian assignment on
wrist distance with a handedness term, and a joint is counted correct when its
distance to the corresponding ground-truth joint is within $5\%$ of the image
diagonal. Keypoints belonging to a ground-truth hand with no matched prediction
count as errors, so failing to render a hand is penalized rather than ignored.

\emph{Motion-Consistency (MC)} measures whether the hands move as they do in
the ground truth using \cite{wang2023videomae}, rather than merely occupying the correct positions at each
instant. It compares the per-frame displacement of matched keypoints between
consecutive frames in the prediction and in the ground truth and reports the
fraction whose motion agrees within the same $5\%$-of-diagonal tolerance used
by PCK. This complements PCK: a model can place a hand correctly in every frame
while moving it along the wrong path, which PCK alone cannot detect, and MC
also captures the drift and jitter that accumulate over a long rollout.

\noindent\textbf{Object metrics.} We follow the matching pipeline of
EgoX~\citep{kang2025egoxegocentricvideogeneration}. Objects are segmented with
SAM~2 and tracked across the clip; each object is described by a DINOv3
embedding of its cropped bounding box, and ground-truth and predicted objects
are matched by one-to-one assignment on embedding cosine similarity subject to
a similarity threshold. For each matched pair we report the \emph{center
location error}, the Euclidean distance between bounding-box centers, and the
\emph{bounding-box IoU}. Location error captures whether the object is placed
correctly on the egocentric canvas, while IoU additionally captures whether its
extent is right; both are reported in normalized image units.

\noindent\textbf{Video metrics.} \emph{FVD} is the Fr\'echet distance between
I3D feature distributions of predictions and ground truth, computed at the
level of a split rather than per video. It captures both the visual
distribution gap and, through the temporal features of I3D, inter-frame
continuity. We additionally report two reference-free VBench dimensions.
\emph{Temporal Flickering (TF)} is the mean absolute difference between
consecutive frames, mapped to $[0,1]$ so that higher is better, and detects
high-frequency temporal noise. \emph{Motion Smoothness (MS)} measures the error
of an off-the-shelf frame-interpolation model~\cite{zhang2025evaluation} in reconstructing the actual
frames from their neighbours, again mapped so that higher is better and thus
penalizing motion that is not temporally coherent. Because a static output can
score well on both, we read them alongside FVD and the object metrics rather
than in isolation.

\subsection{Head tracker for end-to-end streaming}
\label{sec:supp_tracker}

The head tracker of Sec.~\ref{sec:exp_e2e} is causal and built from open-source estimators: Multi-HMR~\citep{baradel2024multihmr} estimates the 3D head position on every second frame, MoGe-2~\citep{wang2025moge2} aligns its metric depth, SemiUHPE~\citep{zhou2025semiuhpe} estimates the head orientation, and a causal One Euro filter~\citep{casiez2012oneeuro} smooths the pose trajectory. A VLM identifies the wearer once at the start.

\noindent\textbf{Latency.} Running the causal head tracker on one additional A100 GPU in parallel with the generator takes $0.91$\,s per $1.375$-s chunk ($33$ frames at $832\times480$ resolution), which is well within the chunk duration and keeps the end-to-end pipelined throughput at $24.3$ FPS. Table~\ref{tab:supp_tracker_latency} details the per-chunk latency of each stage of the tracker.

\begin{table}[h]
  \centering
  \caption{Per-chunk latency of the causal head tracker ($33$ frames, $1.375$\,s of video).}
  \label{tab:supp_tracker_latency}
  \begin{tabular}{lcc}
    \toprule
    Stage & Hardware & Time per chunk \\
    \midrule
    Multi-HMR on 17 of 33 frames (head position) & A100 & $0.68$\,s \\
    SemiUHPE on the head crops (head orientation) & A100 & $0.03$\,s \\
    MoGe-2 on 3 frames (metric depth) & A100 & $0.19$\,s \\
    Wearer tracking, One Euro filter & CPU & $<0.01$\,s \\
    \midrule
    Head tracking in total & -- & $0.91$\,s \\
    \bottomrule
  \end{tabular}
\end{table}

\noindent\textbf{Quality.} The estimated poses deviate from the dataset trajectories by $52.8$\,cm and $22.3^\circ$ on \textsc{New Action}, by $54.5$\,cm and $22.7^\circ$ on \textsc{Unseen} (mostly along the viewing ray, as the wearers stand about $3$\,m from the camera), and by $6.3$\,cm and $10.9^\circ$ on H2O~\citep{h2o} (about $1$\,m). The resulting scores are reported as Ours (est.) in Table~\ref{tab:short_video_comparison}.

\section{Limitations}
\label{sec:supp_limitations}

The vision motivating this work has two parts: streaming Exo2Ego generation,
and the use of that generation to manufacture first-person training data for
embodied agents at scale. We have addressed the former as far as the current
task formulation allows, but the latter remains constrained by a single
assumption that our method---and every existing Exo2Ego method~\citep{luo2024put,liu2024exocentric,mahdi2025exo2egosyn,xu2025egoexogen,kang2025egoxegocentricvideogeneration,park2026egoworld,mahdi2026synchronysequenceexotoegogeneration,wang2026groundedexo2ego} we are aware
of---inherits: the egocentric camera trajectory is given in advance.

This assumption is what restricts the applicability of Exo2Ego to the
comparatively small set of recordings that are paired with tracked first-person
capture, whereas the vast majority of third-person video on the internet has no
associated egocentric trajectory. Removing the assumption is beyond the scope
of this paper, since trajectory estimation is an orthogonal perception problem
rather than a generation one, and since our conditioning and evaluation
apparatus both take calibrated ego intrinsics and extrinsics as input.

We regard the obstacle as temporary rather than fundamental. Head and camera
pose estimation has advanced rapidly, and streaming, causally evaluated models
now exist for this purpose, including egocentric pose estimation models
designed for on-device AR/VR deployment with sub-millisecond latency
\citep{yang2024egoposeformer,li2026egoposeformerv2}, and stateful
reconstruction models that consume an image stream online and maintain a
persistent, continuously updated scene state~\citep{wang2025cut3r}. Once such
estimators are accurate and stable enough to supply a plausible ego trajectory
from in-the-wild third-person footage, they could be composed with our pipeline
at negligible latency. Realizing the full vision is therefore primarily a
matter of that component maturing, rather than of a change to the generative
formulation presented here.

\end{document}